\documentclass[journal]{IEEEtran}
\ifCLASSINFOpdf
\else
\fi

\usepackage{color, soul}
\soulregister\cite7 
\soulregister\citep7 
\soulregister\citet7 
\soulregister\ref7 
\soulregister\pageref7 

\usepackage{amsmath, graphicx, multirow, subfig}

\begin{document}
	\def \I{ \mathbf{I} }
	\def \H{ \mathbf{H} }
	\def \n{ \mathbf{n} }
	\def \u{ \mathbf{u} }

	%
	\title{Hierarchical Similarity Learning for Aliasing Suppression Image Super-Resolution}
	%
	%
	%
	
	\author{
		Yuqing~Liu,
		Qi~Jia,
		Jian~Zhang,
		Xin Fan,~\IEEEmembership{Senior Member,~IEEE,}
		Shanshe Wang,
		Siwei~Ma,~\IEEEmembership{Member,~IEEE,}
		and~Wen~Gao,~\IEEEmembership{Fellow,~IEEE}
		\thanks{Y. Liu is with the School of Software, Dalian University of Technology,
			Dalian 116620, China (e-mail:liuyuqing@mail.dlut.edu.cn).}
		\thanks{Q. Jia and X. Fan are with International School of Information Science and Engineering, Dalian University of Technology, Dalian 116620, China (e-mail:
			jiaqi@dlut.edu.cn; xin.fan@dlut.edu.cn).}
		\thanks{J. Zhang is with the School of Electronic and Computer Engineering, Peking University Shenzhen Graduate School, Shenzhen, China. (e-mail: zhangjian.sz@pku.edu.cn)}
		\thanks{S. Wang, S. Ma, and W. Gao are with the School of Electronics Engineering and Computer Science, Institute of Digital Media, Peking University, Beijing 100871, China (e-mail: sswang@pku.edu.cn; swma@pku.edu.cn; wgao@pku.edu.cn).}
	}
	
	%
	%

	\markboth{Manuscript submitted to IEEE Transactions on Neural Networks and Learning Systems}%
	{Shell \MakeLowercase{\textit{et al.}}: Bare Demo of IEEEtran.cls for IEEE Journals}
	%



	\maketitle
	
	\begin{abstract}
		As a highly ill-posed issue, single image super-resolution (SISR) has been widely investigated in recent years. The main task of SISR is to recover the information loss caused by the degradation procedure. According to the Nyquist sampling theory, the degradation leads to aliasing effect and makes it hard to restore the correct textures from low-resolution (LR) images. In practice, there are correlations and self-similarities among the adjacent patches in the natural images. This paper considers the self-similarity and proposes a hierarchical image super-resolution network (HSRNet) to suppress the influence of aliasing. We consider the SISR issue in the optimization perspective, and propose an iterative solution pattern based on the half-quadratic splitting (HQS) method. To explore the texture with local image prior, we design a hierarchical exploration block (HEB) and progressive increase the receptive field. Furthermore, multi-level spatial attention (MSA) is devised to obtain the relations of adjacent feature and enhance the high-frequency information, which acts as a crucial role for visual experience. Experimental result shows HSRNet achieves better quantitative and visual performance than other works, and remits the aliasing more effectively. 
	\end{abstract}
	

	%
	\IEEEpeerreviewmaketitle

	\section{Introduction}
	\label{sec:introduction}
	%
	%
	%
	%
	\IEEEPARstart{I}{mage} super-resolution is a classical issue in image processing area~\cite{sr_review_pami2020}. Given a low-resolution (LR) image, the task of single image super-resolution (SISR) is to find the corresponding high-resolution (HR) instance with refined details. SISR has been considered in different applications, such as video/image compression, object detection and semantic segmentation.
	
\begin{figure}[t]
		\captionsetup[subfloat]{labelformat=empty, justification=centering}
		\begin{center}
			\newcommand{\rowArg}{1.7cm}
			\newcommand{\fullSize}{4.6cm}
			\newcommand{\fullWidth}{2.5cm}
			\newcommand{\patchSize}{1.9cm}
			\scriptsize
			\setlength\tabcolsep{0.05cm}
			\begin{tabular}[b]{c c c c c}
				\multicolumn{2}{c}{\multirow{2}{*}[\rowArg]{
						\subfloat[Example image from Urban100~\cite{urban100}]{\includegraphics[height=\fullSize, width=\fullWidth]{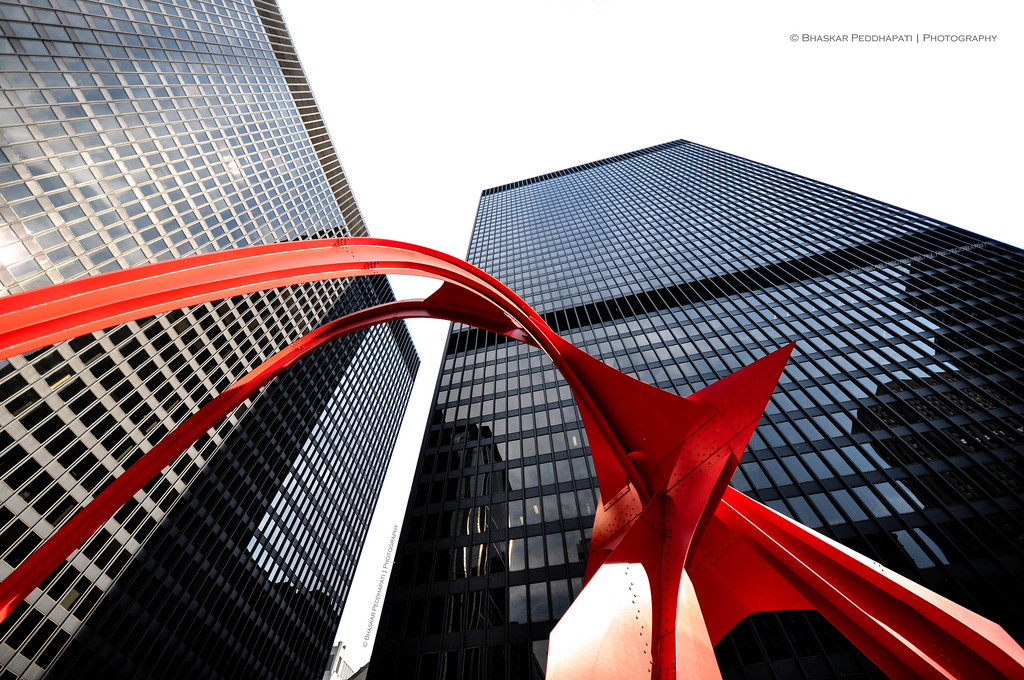}}}} &
				\subfloat[HR~\protect\linebreak(PSNR/SSIM)]{\includegraphics[width = \patchSize, height = \patchSize]{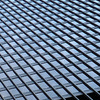}} &
				\subfloat[LR~\protect\linebreak(18.24/0.6158)]{\includegraphics[width = \patchSize, height = \patchSize]{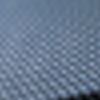}} &
				\subfloat[LapSRN~\cite{lapsrn_pami2019}~\protect\linebreak(19.54/0.7614)]{\includegraphics[width = \patchSize, height = \patchSize]{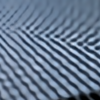}} \\
				& &			
				\subfloat[VDSR~\cite{vdsr_cvpr2016}~\protect\linebreak(19.49/0.7584)]{\includegraphics[width = \patchSize, height = \patchSize]{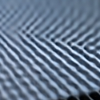}} &
				\subfloat[MSRN~\cite{msrn_eccv2018}~\protect\linebreak(20.28/0.8170)]{\includegraphics[width = \patchSize, height = \patchSize]{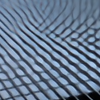}} &
				\subfloat[HSRNet~\protect\linebreak(\textbf{20.84/0.8285})]{\includegraphics[width = \patchSize, height = \patchSize]{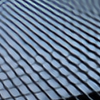}} \\
			\end{tabular}
		\end{center}
		\setlength{\abovecaptionskip}{0pt plus 2pt minus 2pt}
		\setlength{\belowcaptionskip}{0pt plus 2pt minus 2pt}
		\vspace{-0.2cm}
		\caption{Visual quality comparison for different SISR methods.}
		\label{fig:slogan}
	\end{figure}

	As a highly ill-posed issue, there is numerous information loss caused by the degradation. According to the Nyquist sampling theory~\cite{nyquist_tit2009}, the degradation procedure decreases the signal frequency and leads to aliasing, which makes it hard to recover the correct textures. An example of this effect is shown in Figure \ref{fig:slogan}. The higher positions of the building contain plentiful high frequency information and suffer severer aliasing after degradation. The texture mixture makes it hard to find the accurate structural textures. 
	
	Fortunately, there are correlations and self-similarities among the adjacent patches in the natural images, and the frequency change in natural image is continuous and smooth. In this point of view, the adjacent patches with lower frequency information is helpful to restore the high frequency information and suppress the aliasing. Figure \ref{fig:slogan} provides an example of this hypothesis. The textures in the building are similar. From the bottom to the top, the lines becomes dense, and the frequency gets higher. As such, a refined local image prior can be a good guidance to restore the high-frequency areas.
	
	Recently, convolutional neural network (CNN) has proved to be an effective design for exploring the inherent correlations of features and restoring the missing information. SRCNN~\cite{srcnn_pami2016} is the first CNN-based work for SISR problem. After SRCNN, VDSR~\cite{vdsr_cvpr2016}, EDSR~\cite{edsr_cvprw2017}, LapSRN~\cite{lapsrn_pami2019}, and other works with elaborate designs builds the network deeper and wider to boost the network representation and restoration capacity. These works focus on elaborate block designs and utilize straightforward networks to recover the missing textures, which seldom concentrate on the local image prior.
	
	There are also CNN-based works recovering the degraded images with the help of external information. Reference-based image super-resolution (RefSR) is also a feasible methodology to restore the degraded images, which requires an extra image as prior guidance for texture recovery. Traditional example-based methods have been considered for image restoration~\cite{landmarksr_tip2013, anchorsr_iccv2013, gssr_tmm2017} with good visualization performance. There are also deep learning based works for effective texture transfer~\cite{crossnet_eccv2018, srntt_cvpr2019}. Although the reference can give an outstanding prior for recovering the missing information, there are two critical issues for RefSR to restore the correct textures. On one hand, the quality of reference image directly influences the performance of restoration. On the other hand, different from the self-similarity in nature image, the provided reference image contains diverse information for restoration, and leads to excrescent and incorrect textures.
	
	In this paper, we aim to find a formula solution for the SISR problem in the optimization perspective, and restore the missing information with the help of similarity learning. 
	Specifically, we design an iterative SISR network based on the half-quadratic splitting (HQS) strategy.
	To consider the local image prior, a hierarchical exploration block (HEB) is proposed to explore the texture feature.
	The hierarchical exploration in HEB progressively recovers the missing texture by filters with incremental receptive fields and gradually enlarges the local information prior.
	Besides the hierarchical exploration, a multi-level spatial attention (MSA) is designed to address the adjacent spatial correlation of feature and concentrate more on high-frequency information. Based on the optimization-based network backbone and the advanced block designs, we design an hierarchical image super-resolution network termed as HSRNet. Experimental results show our HSRNet achieves better quantitative and visual performance than other image SR works. Specially, HSRNet can remit the aliasing more effectively. Figure~\ref{fig:slogan} shows an example visual quality comparison for different SISR methods. In the figure, existing methods~\cite{lapsrn_pami2019, vdsr_cvpr2016, msrn_eccv2018} recover the correct grids at the lower positions, but fail on the top of the batch. In contrast, HSRNet can recover more accurate lines and textures under heavy aliasing.
	
	Our contributions are summarized as follows:
	\begin{itemize}
		\item We consider the image super-resolution in the optimization perspective, and design an end-to-end SISR network for image restoration based on the half-quadratic splitting (HQS) strategy.
		
		\item We consider the self-similarity to suppress the aliasing. To fully utilize the local image prior, we design the hierarchical exploration block (HEB) and the multi-level spatial attention (MSA), and devise a hierarchical image super-resolution network (HSRNet).
		
		\item Experimental results show our HSRNet achieves better quantitative and visual performance than other image super-resolution works. Specially, HSRNet can remit the aliasing more effectively and recover more accurate structural textures.
	\end{itemize}
	
	\section{Literature Review}
	
	\subsection{Deep learning for Single Image Super-Resolution}
	Convolutional neural network (CNN) has been widely utilized in various computer vision tasks and demonstrates its powerful performance. The first CNN-based work is SRCNN~\cite{srcnn_pami2016} with a three-layer neural network, which followed a sparse-coding like structure. After SRCNN, deeper and wider CNNs were proposed for boosting the performance. VDSR~\cite{vdsr_cvpr2016} built a very deep network with 20 convolutional layers for better restoration capacity. In ESPCN~\cite{espcn_cvpr2016}, sub-pixel convolution was firstly introduced for up-sampling, and has been widely used in later works. EDSR~\cite{edsr_cvprw2017} pointed out that batch normalization is not suitable for SISR problem, and built a very deep network with residual blocks. SRDenseNet~\cite{srdensenet_iccv2017}, inspired by densely connection, also achieved state-ot-the-art performance with well-designed network architecture. RDN~\cite{rdn_cvpr2018} combined the residual connection and densely connection, and proposed a residual dense block for effective image restoration. RCAN~\cite{rcan_eccv2018} considered residual-in-residual design for effective information and gradient transmission. Inspired by Laplacian Pyramid, LapSRN~\cite{lapsrn_pami2019} firstly proposed a progressive Laplacian network for multiple scale image super-resolution. OISR~\cite{oisr_cvpr2019} considered an ODE-Inspired network and achieved good performance. DBPN~\cite{dbpn_pami2020} obtained the back projection method for SISR and designed a network to restore the image. SRFBN~\cite{srfbn_cvpr2019} also considered feedback and iteration for effective image restoration with restricted parameters. Recently, DAN~\cite{dan_cvpr2020}, CS-NL~\cite{csnl_cvpr2020}, and other works also utilize elaborate network designs and achieve amazing restoration performance.
	
	There are also lightweight works for efficient SISR. CARN~\cite{carn_eccv2018} utilized a cascading recursive network to restore the image with fast speed. IDN~\cite{idn_cvpr2018} considered information distillation mechanism into image super-resolution and successfully decreased the computation complexity and network parameters. To further investigate the effectiveness of information distillation, IMDN~\cite{imdn_mm2019} proposed the multi-distillation strategy and achieved state-of-the-art performance. RFDN~\cite{aim2020} improved the architecture of IMDN and further boosted the restoration capacity.
	
	Besides effective block designs, attention mechanism has also become an important part for SISR network, which aims to concentrate more on important information and area. As far as we know, channel-wise attention block was firstly proposed in SENet~\cite{senet_cvpr2018} to boost the network representation, and then was utilized in RCAN~\cite{rcan_eccv2018} to concentrate more on important feature channels. RFDN~\cite{aim2020} considered an enhanced spatial attention for comprehensive correlation learning. Besides local attentions, CS-NL~\cite{csnl_cvpr2020} proposed a cross-scale non-local attention mechanism to achieve the state-of-the-art performance.
	
	\subsection{Image Super-Resolution in the Optimization View}
	There are also works considering the problem in the optimization perspective. IRCNN~\cite{ircnn_cvpr2017} utilized half-quadratic splitting (HQS) strategy and built a network to learn the denoiser prior for image restoration. DPSR~\cite{dpsr_cvpr2019} proposed a novel degradation model and used HQS to find the closed-form solution. USRNet~\cite{usrnet_cvpr2020} combined the model-based and learning-based methodologies and proposed a deep unfolding super-resolution network for non-blind SISR. Liu \textit{et al.} designed ISRN~\cite{isrn_tmm2021} to perform end-to-end image super-resolution inspired by HQS.
	
	Besides HQS, there are other strategies to model the SISR issue and find the solution. Plug-and-Play ADMM~\cite{plug_and_play_admm_tci2017} has been proposed for image restoration with a plug-and-play denoiser. Deng~\textit{et al.} proposed a deep coupled ISTA network for multi-model image super-resolution~\cite{jmdl_tip2020}. ISTA-Net~\cite{istanet_cvpr2018} considered an unfolding network based on the ISTA algorithm, and achieved good performance on compressive sensing and image restoration. ADMM-Net~\cite{admmnet_iccv2019} also proposed an ADMM-based network for reconstructing the compressive sensing images.
	
	\begin{figure*}[t]
		\centering
		\includegraphics[width=\linewidth]{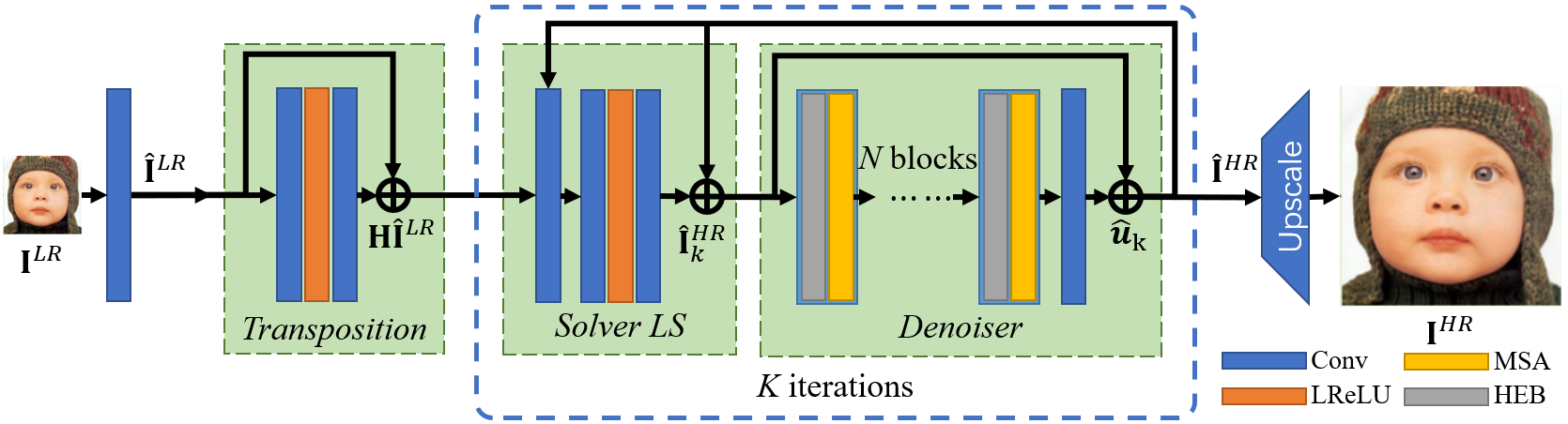}
		\caption{Network structure of HSRNet. The space transformation is conducted by convolution and upscale. Three components are proposed to perform the optimization}
		\label{fig:network}
	\end{figure*}
	\section{Methodology}
	In this section, we will introduce our HSRNet in the following manner. Firstly we will analyze the issue according to HQS strategy, and describe the network design of our HSRNet. Then, the detail of hierarchical exploration block (HEB) and multi-level spatial attention (MSA) will be discussed. Finally, we will talk about the difference between HSRNet and other related works.
	
	\subsection{Network Design}
	The observation model of SISR issue is,
	\begin{equation}
		\label{eq:1}
		\mathbf{I}^{LR}=\mathbf{H}\mathbf{I}^{HR}+\mathbf{n},
	\end{equation}
	where $\mathbf{I}^{LR}$, $\mathbf{I}^{HR}$ are corresponding LR and HR instances, $\mathbf{H}$ is the degradation matrix and $\mathbf{n}$ is the noise term. To find $\mathbf{I}^{HR}$ from $\mathbf{I}^{LR}$, based on the observation model, the optimization target of SISR can be written as,
	\begin{equation}
		\label{eq:2}
		\mathbf{I}^{HR}=\arg\min_{\mathbf{I}^{HR}}\frac{1}{2}||\mathbf{I}^{LR}-\mathbf{HI}^{HR}||_\ell^2+\Phi(\mathbf{I}^{HR}),
	\end{equation}
	where $\Phi(\cdot)$ denotes the prior term for HR images.
	
	Let $\mathbf{u}=\mathbf{I}^{HR}$, according to the HQS strategy, the optimization target can be cast into the iterative solution, as,
	\begin{equation}
		\label{eq:3}
		\left\lbrace 
		\begin{aligned}
			\I^{HR}_{k+1} 	&= \arg\min_{\I^{HR}} \frac{1}{2}||\I^{LR}-\H\I^{HR}||_\ell^2 + \frac{\beta_k}{2}||\I^{HR}-\u_k||^2_\ell, \\
			\u_{k+1}		&= \arg\min_\u \frac{\beta_k}{2}||\I^{HR}_{k+1}-\u||^2_\ell + \Phi(\u),		
		\end{aligned}
		\right. 
	\end{equation}
	where $\beta$ is a constant of penalty term, and increases with the accumulation of iteration times. It should be addressed that the solution of $\u_{k+1}$ has the same situation of a denoising problem, and $\I^{HR}_{k+1}$ has a closed-form least square (LS) solution, where,
	\begin{equation}
		\label{eq:4}
		\I^{HR}_{k+1} = (\H\H^\mathrm{T}+\beta\I)^{-1}(\H^\mathrm{T}\I^{LR}+\beta\u_{k}).
	\end{equation}
	
	In fact, when $\H$ is hard to be found, or even without explicit mathematical expression, it is challenging to find the solution of Equation (\ref{eq:4}). A possible strategy is to learn the hidden degradation from training data pairs. In this point of view, Equation (\ref{eq:4}) should be regarded as a CNN-based function with the input as $\H^\mathrm{T}\I^{LR}$ and $\u_{k}$, and the output as $\I^{HR}_{k+1}$. Similarly, the solution of $\u_{k+1}$ can be also considered as a denoiser with the input $\I^{HR}_{k+1}$. As such, Equation (\ref{eq:3}) can be converted into an iterative CNN-based structure for finding the solution, which can learn the hyper-parameters more effectively.
	
	However, there is a critical issue that both $\u$ and $\I^{HR}_{k}$ are with high resolution, which cause higher computation complexity. In this point of view, a specific space with low resolution should be considered to process the HR images and features. To make the optimization in the low resolution space, we devise the HSRNet to convert the spaces and perform the optimization iteratively.
	
	Figure~\ref{fig:network} shows the network structure of HSRNet. Firstly, a convolution converts the input LR image into a specific space, as,
	\begin{equation}
		\label{eq:5}
		\hat{\I}^{LR} = Conv(\I^{LR}),
	\end{equation}
	where $\hat{\cdot}$ means the representation in the specific space.
	
	After space transformation, \textit{Transposition} module aims to learn the $\H^\mathrm{T}$ and performs it on $\hat{\I}^{LR}$, such that,
	\begin{equation}
		\label{eq:6}
		\H^\mathrm{T}\hat{\I}^{LR}=Trans(\hat{\I}^{LR}),
	\end{equation}
	where $Trans(\cdot)$ is the \textit{Transposition} module. As shown in the figure, this module is composed of two convolutional layers and a Leaky ReLU activation.
	
	With the learned $\H^\mathrm{T}\hat{\I}^{LR}$, \textit{Solver LS} and \textit{Denoiser} are mainly designed to generate the solution of Equation~(\ref{eq:3}) separately. HSRNet performs the iteration for several times. For the $k$-th iteration, there are,
	\begin{equation}
		\label{eq:7}
		\hat{\I}^{HR}_{k+1} = SolverLS(\H^\mathrm{T}\hat{\I}^{LR}, \hat{\u}_{k}),
	\end{equation}
	and
	\begin{equation}
		\label{eq:8}
		\hat{\u}_{k+1} = Denoiser(\hat{\I}^{HR}_{k+1}).
	\end{equation}
	The \textit{Solver LS} aims to calculate the least square solution in the specific space by neural network, which is composed of three convolutional layers and a Leaky ReLU. The \textit{Denoiser} is composed of MSAs and HEBs and a convoluitonal layer, which are sequentially stacked. 
	
	Finally, the HR image is restored from the specific space after $K$-th iteration, as,
	\begin{equation}
		\label{eq:9}
		\I^{HR} = Upscale(\hat{\I}^{HR}).
	\end{equation}
	The \textit{Upscale} module is simply composed of one convolutional layer and a sub-pixel convolution.
	
	In the next sections, we will introduce the designed hierarchical exploration block (HEB) and the multi-level spatial attention (MSA) for HSRNet. The proposed components concentrate on the hierarchical correlations among features and the local image prior for effective aliasing suppression.
	
	\subsection{Hierarchical Exploration Block}
	\label{sec:heb}
	\begin{figure}[t]
		\centering
		\includegraphics[width=\linewidth]{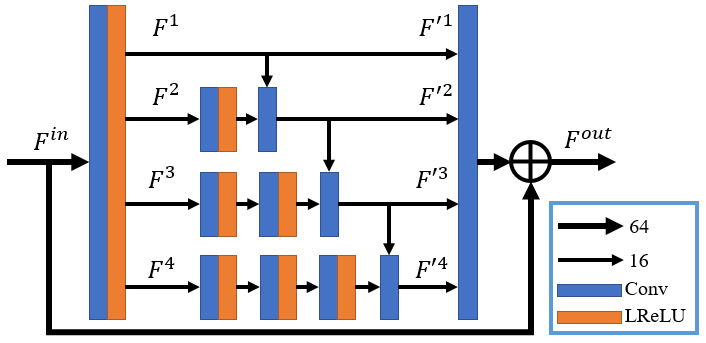}
		\caption{Structure of hierarchical exploration block (HEB).}
		\label{fig:heb}
	\end{figure}
	
	Figure \ref{fig:heb} shows the design of HEB. Let $F^{in}$, $F^{out}$ denote the input and output features separately. One convolution with LeakyReLU firstly explore the input feature, as,
	\begin{equation}
		\label{eq:10}
		F^{feat}=f(F^{in}),
	\end{equation}
	where $f(\cdot)$ denotes the feature exploration operation by one convolution with LeakyReLU.
	
	After exploration, the feature $F^{feat}$ is separated into four features with equal channel number. This separation is described as,
	\begin{equation}
		\label{eq:11}
		[F^1, F^2, F^3, F^4] = F^{feat},
	\end{equation}
	where $F^{feat}$ is with 64 channels, while $\{F^i\}_{i=1}^4$ are with 16 channels.
	
	As shown in Figure~\ref{fig:heb}, the four separated features are processed with different receptive fields for further hierarchical information aggregation. For $F^1$, there is no operation and the identical information is preserved. The processed feature is considered as,
	\begin{equation}
		\label{eq:12}
		F'^1=F^1.
	\end{equation}
	For the $i$-th feature $F^i$ where $i\geq2$, it will be firstly explored for $(i-1)$ times to increase the receptive field, and then combined with $F'^{i-1}$ to consider the neighborhood information. This exploration can be described as,
	\begin{equation}
		\label{eq:13}
		F'^i=Conv([f^{i-1}(F^i), F'^{i-1}]),
	\end{equation}
	where $f^{i-1}$ means perform the exploration for $(i-1)$ times.
	
	Finally, the hierarchical features will be concatenated for jointly consideration. The output feature of HEB is,
	\begin{equation}
		\label{eq:14}
		F^{out} = F^{in} + Conv([F'^1, F'^2, F'^3, F'^4]).
	\end{equation}
	
	\begin{figure}[t]
		\centering		
		\subfloat[Laplacian Pyramid]{\includegraphics[width = 0.448\linewidth]{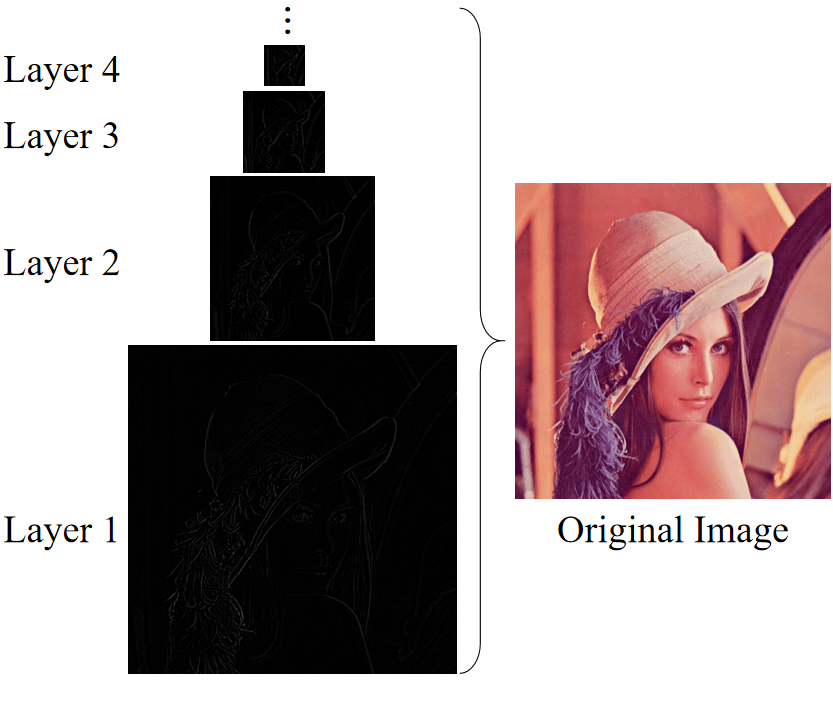}}
		\subfloat[Hierarchical Exploration in HEB]{\includegraphics[width = 0.551\linewidth]{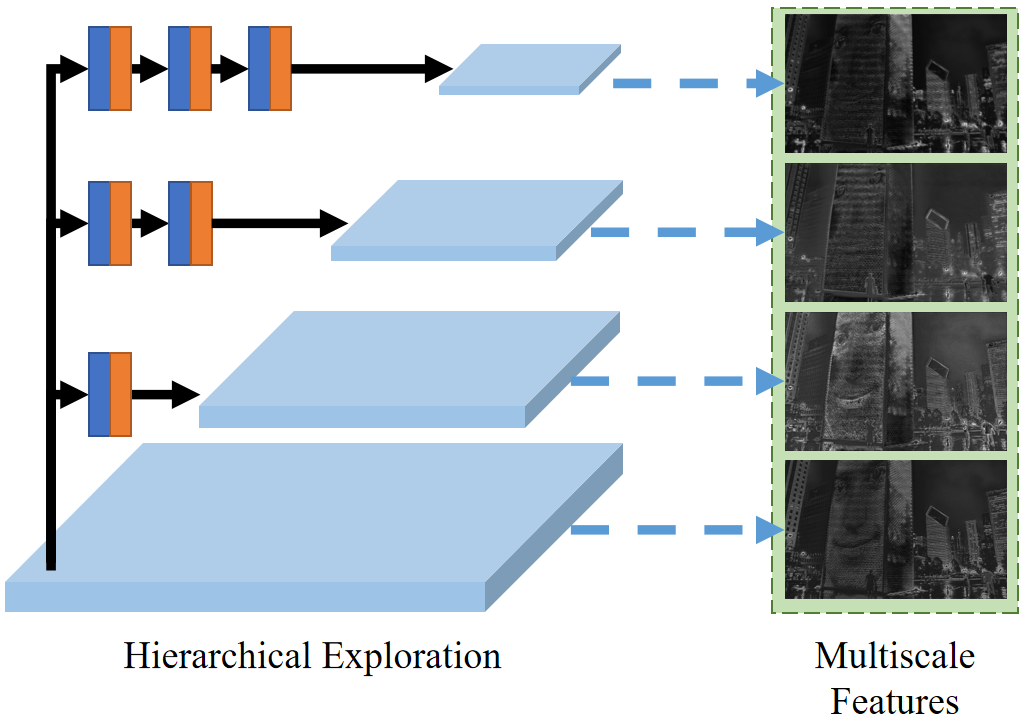}}
		\caption{Comparison between the hierarchical exploration in HEB and the Laplacian Pyramid.}
		\label{fig:heb-lap}
	\end{figure}

	The HEB obtains the local image prior in a progressive manner. With the increase of feature index $i$, the receptive field becomes larger and more adjacent information is considered to restore the mixed textures. The hierarchical exploration strategy in HEB has a similar structure to the Laplacian Pyramid. Figure~\ref{fig:heb-lap} shows the comparison between the hierarchical exploration in HEB and the Laplacian Pyramid. 
	
	{
		With the increase of pyramid layer, the Laplacian Pyramid focuses more on the global information. This is in accordance with the hierarchical exploration in the HEB. In the Figure~\ref{fig:heb-lap}~(a) we can find that the Laplacian Pyramid separates the original image into several layers. The first layer contains the most details information of the image. With the increase of the layer, the residual maps have higher responses on the global information. Similarly, we conduct the hierarchical exploration in HEB by stacking different numbers of the convolutional layers. With the increase of convolutional layers, the receptive field becomes larger, and the structural and larger scale information is gradually addressed.
	}
	
	{
		Different from the Laplacian Pyramid that downsamples the image by interpolation and Gaussian blur, the HEB utilizes convolutional layers to increase the receptive field without changing the resolution of the feature map. HEB can adaptively learn the downscaling parameters in the training step, and produces clearer and suitable features for restoration. It should be noticed there is a concatenation operation between features from different scales in the HEB. With the help of the multi-scale information, structural information from adjacent patches is gradually addressed to consider the local similarity and recover the missing textures.	
	}
	
	\subsection{Multi-level Spatial Attention}
	\begin{figure}[t]
		\centering
		\includegraphics[width=\linewidth]{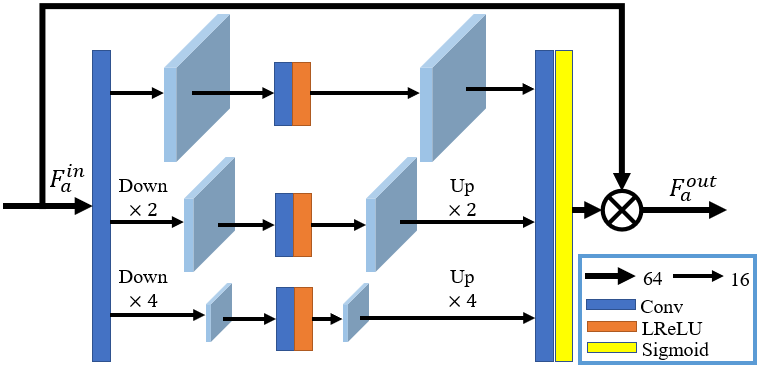}
		\caption{Structure of multi-level spatial attention (MSA).}
		\label{fig:msa}
	\end{figure}
	Besides the feature exploration, attention mechanism is devised to concentrate more on the important area and complex information. Let $F^{in}_a$, $F^{out}_a$ denote the input and output features of MSA separately. The input feature is firstly processed and separated into three features with equal channel number, which is described as,
	\begin{equation}
		\label{eq:15}
		[F_a^1, F_a^2, F_a^3] = Conv(F^{in}_a),
	\end{equation}
	where $F_a^1$, $F_a^2$, and $F_a^3$ are with 16 channels, and $F^{in}_a$ is with 64 channels. 
	
	After separation, max-pooling is performed on features with different scaling factors to obtain the multi-level spatial representation. One convolution with LeakyReLU is utilized to explore the multi-level feature. Then, bilinear interpolation is utilized to upscale the features, and keep the shapes before and after multi-level exploration. This operation can be described as,
	\begin{equation}
		\label{eq:16}
		F'^1_a = f(F^1_a),
	\end{equation}
	\begin{equation}
		\label{eq:17}
		F'^2_a = (f((F^2_a)\downarrow_{\times2}))\uparrow_{\times2},
	\end{equation}
	\begin{equation}
		\label{eq:18}
		F'^3_a = (f((F^3_a)\downarrow_{\times4}))\uparrow_{\times4},
	\end{equation}
	where $(\cdot)\downarrow$ means the max pooling, $(\cdot)\uparrow$ means the bilinear upsampling, and $f(\cdot)$ denotes the feature exploration operation by one convolution and a Leaky ReLU.
	
	After multi-level exploration, the spatial features are aggregated and processed to obtain the attention map. Finally, the output of MSA is,
	\begin{equation}
		\label{eq:19}
		F_a^{out} = \sigma(Conv([F'^1_a, F'^2_a, F'^3_a])),
	\end{equation}
	where $\sigma(\cdot)$ is the Sigmoid activation.

	\begin{figure*}[t]
		\centering
		\includegraphics[width=0.9\linewidth]{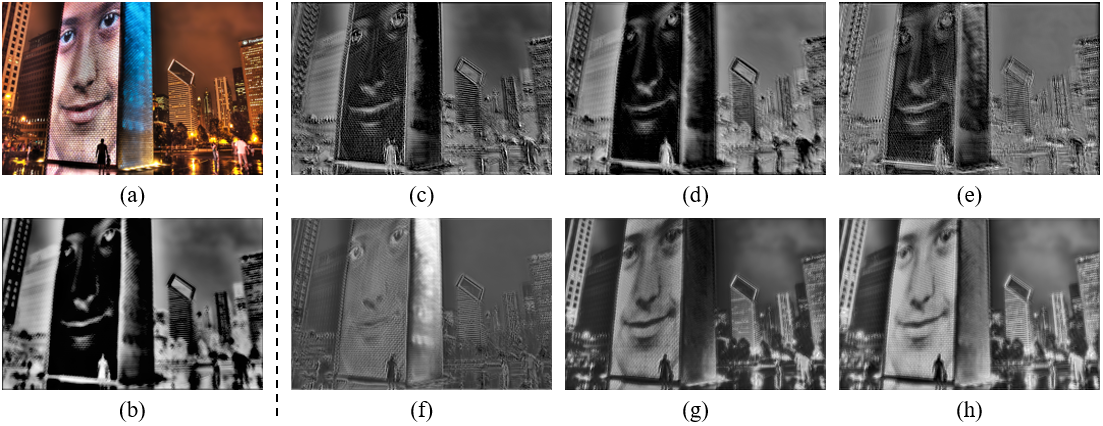}
		\caption{Visualized feature maps of hierarchical exploration. (a): input image. (b): $F^1$ in Eq. (\ref{eq:12}). (c)-(e): the processed features with different receptive fields, which denote $f^1(F^2)$ to $f^3(F^4)$ in Eq.(\ref{eq:13}) separately. (f)-(h): the explored features with different scales, which denote $F'^2$ to $F'^4$ in Eq.(\ref{eq:13}) separately. According to the visualization results, the structural and high-frequency information is gradually addressed, which is in accordance with the local image prior for feature exploration.}
		\label{fig:vis-heb}
	\end{figure*}
	\section{Experiment}
	\subsection{Settings}
	In HSRNet, the convolutional layers in \textit{Transposition} and \textit{Solver LS} are set with 64 filters. There are $N=10$ blocks in the \textit{Denoiser} module, and $K=3$ iterations for optimization. We train the network for 1000 epochs by Adam optimizer with learning rate $lr=10^{-4}$. The loss function is chosen as L1-loss.
	
	We use DIV2K~\cite{div2k} dataset to train the network. DIV2K is a high resolution dataset for image SR, which contains 900 instances with 2K resolution. We use the peak signal-to-noise ratio (PSNR) and structural similarity (SSIM) to measure the restoration performance. Five testing benchmarks are chosen to compare the performance: Set5~\cite{set5}, Set14~\cite{set14}, BSD100~\cite{b100}, Urban100~\cite{urban100} and Manga109~\cite{manga109}. The patch size is chosen as $48\times48$ for the LR input, and the scaling factors are chosen as $\times2$, $\times3$ and $\times4$. The degradation operation is chosen as bicubic down (\textbf{BI}).

	\subsection{Model Analysis}
	
	\subsubsection{{Investigation on the Optimization Modules}}
	
	\begin{figure}[t]
		\centering
		\includegraphics[width = \linewidth]{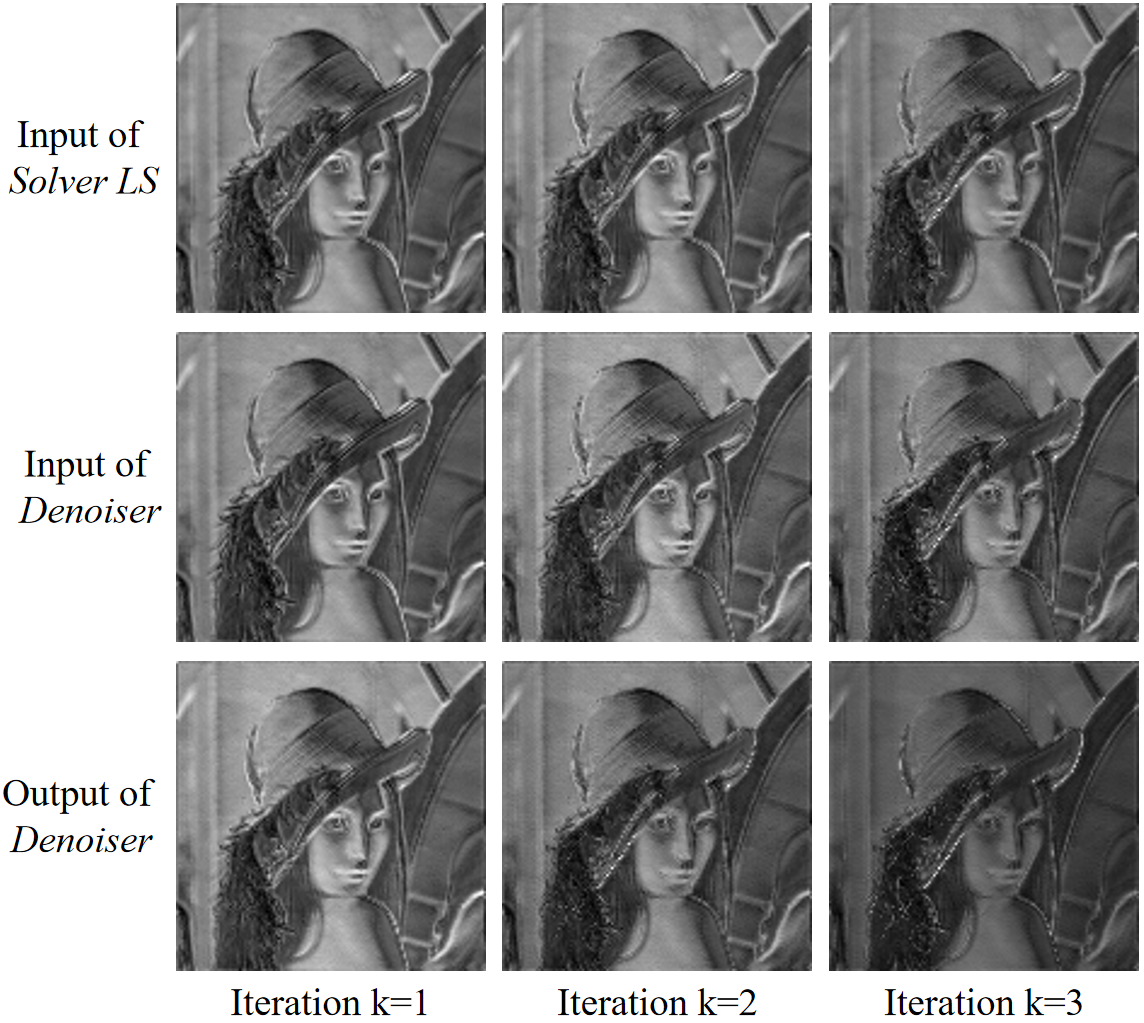}
		\caption{{Visualized feature maps of the input of the \textit{Solver LS}, the input of the \textit{Denoiser}, and the output of the \textit{Denoiser} from different iterations. Zoom up for better view.}}
		\label{fig:r1q1}
	\end{figure}
	
	{
		 In this section, we verify the functions of the proposed modules with showing the feature maps. The feature maps of the input of the \textit{Solver LS}, the input of the \textit{Denoiser}, and the output of the \textit{Denoiser} are shown in Figure~\ref{fig:r1q1}. We choose the maximum values of different channels in the feature maps for visualization, which are normalized in range $0$ to $1$. It should be noticed that the input of the \textit{Denoiser} is just the output of the \textit{Solver LS}. In this point of view, every three feature maps in the same column are produced sequentially by the modules. Every three feature maps in the same row are produced by the same module from different iterations.
	}
	
	{
		For each row of the figure, we can find that the feature maps become clearer with the increase of the iteration. This is in accordance with the optimization scheme. For each column of the figure, the features are sequentially processed by the \textit{Solver LS} and the \textit{Denoiser}. We can find that the \textit{Solver LS} reduces the artifacts and jaggies of the edges. The jaggies are introduced by the $\H^\mathrm{T}\hat{\I}^{LR}$ in the Equation~(\ref{eq:7}), which contains the low-resolution information. \textit{Solver LS} restores the high-resolution information and preserves the fidelity for the input feature maps, which can be regarded as a feasible solution dominated by the $\H^\mathrm{T}\hat{\I}^{LR}$ and the $\hat{\u}$. The \textit{Denoiser} further suppresses the response of the low-frequency areas and the jaggies, and enhances the edges and lines. The structures at the back of the human become sharper after the \textit{Denoiser}. Besides, the responses of the edges of the hat and the human become higher after the \textit{Denoiser}. As such, the \textit{Denoiser} finds a feasible solution of $\hat{\u}$ with higher responses on the structural information.
	}
	
	\begin{figure}[t]
		\captionsetup[subfloat]{labelformat=empty, justification=centering}
		\begin{center}
			\subfloat{\includegraphics[width = 0.32\linewidth]{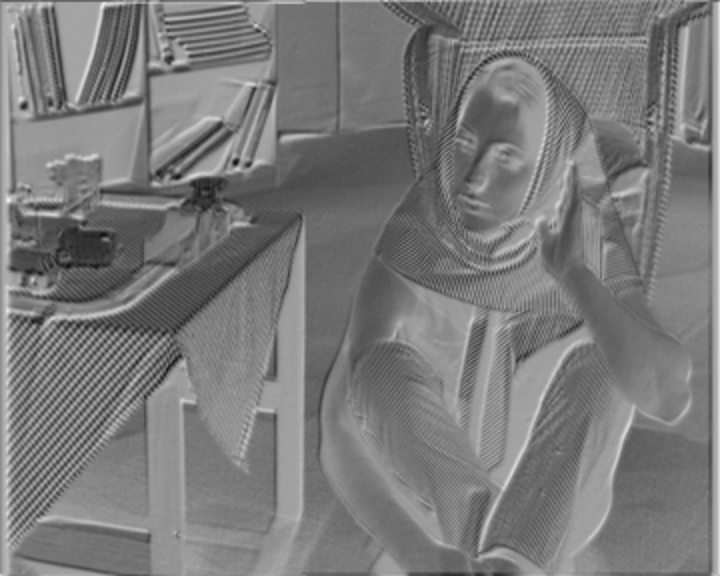}}
			\subfloat{\includegraphics[width = 0.32\linewidth]{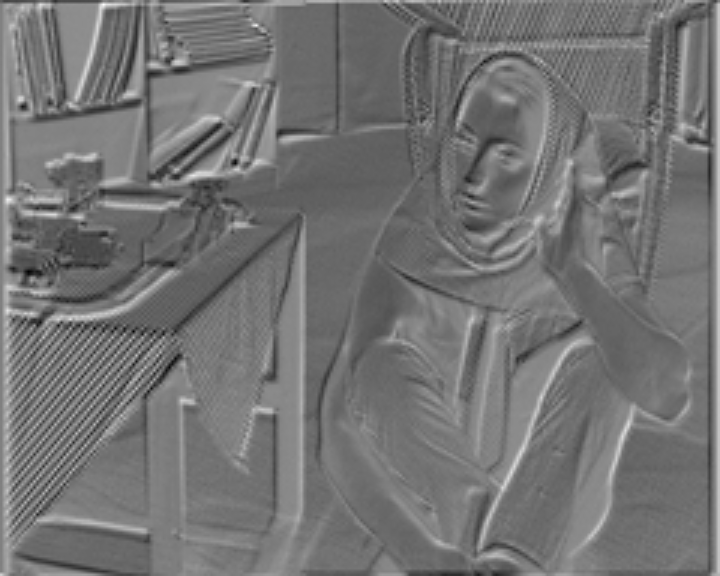}}
			\subfloat{\includegraphics[width = 0.32\linewidth]{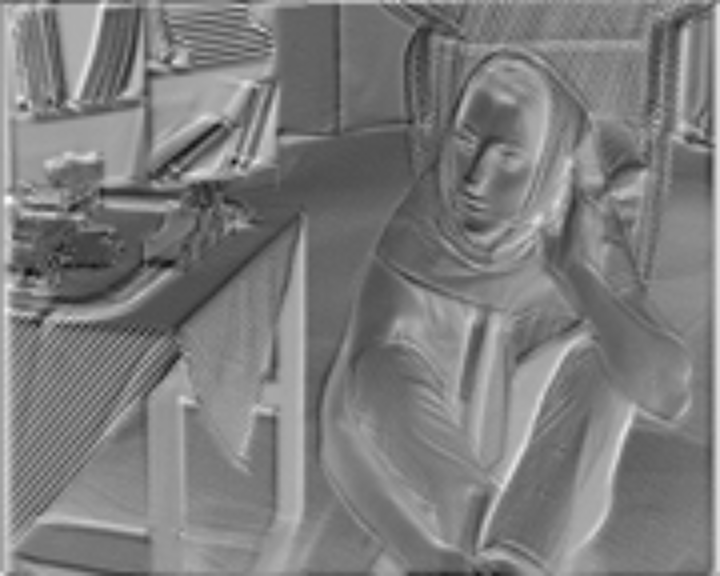}}
			
			\subfloat[(a) Scaling Factor $\times2$]{\includegraphics[width = 0.32\linewidth]{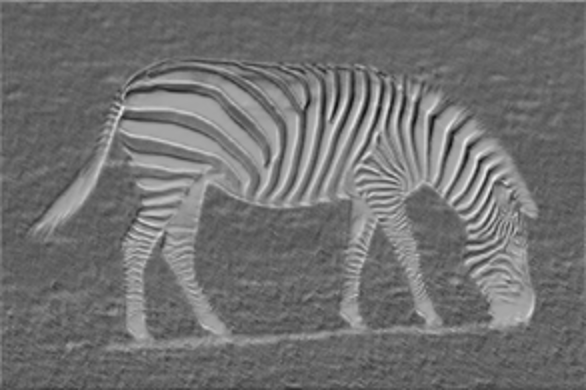}}
			\subfloat[(b) Scaling Factor $\times3$]{\includegraphics[width = 0.32\linewidth]{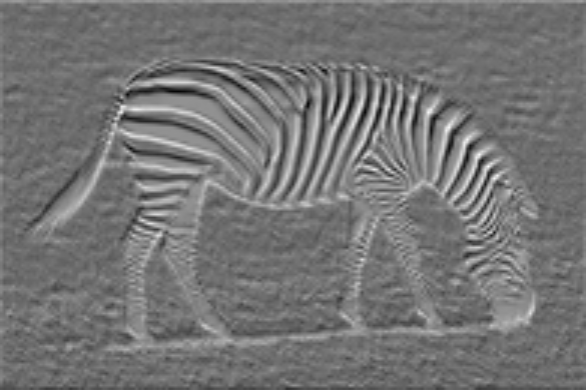}}
			\subfloat[(c) Scaling Factor $\times4$]{\includegraphics[width = 0.32\linewidth]{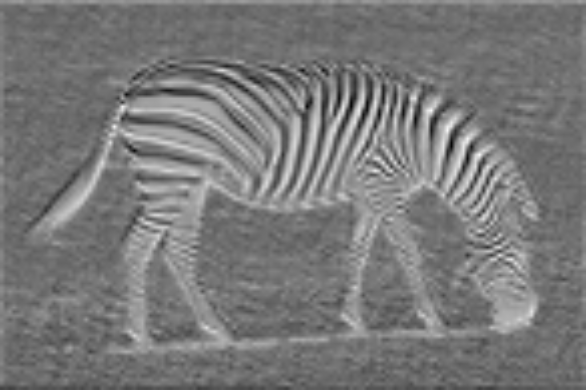}}
		\end{center}
		\caption{{Visualized feature maps of $\hat{\I}^{LR}$ from different scales.}}
		\label{fig:r1q2}
	\end{figure}
	
	{
		There is one convolutional layer at the begin of the network. The convolution converts the image into feature maps with the same size of $\I^{LR}$. By using this conversion, we can effectively reduce the computation cost. With the increase of scaling factors, the resolution of LR images becomes smaller, and the information is limited.
	}
	
	{
		To analyze the effectiveness of different scaling factors, we compare the feature maps from differnet LR inputs and restore them to the same resolution. The visualized feature maps are $\hat{\I}^{LR}$ in the Equation~(\ref{eq:5}). We calculate the average response of different channels and normalize them in range 0 to 1. The visualized feature maps with different scaling factors are shown in the Figure~\ref{fig:r1q2}.
	}
	
	{
		In the figure, we can find that the features become blurry and the information is lost with the increase of scaling factors. The textures of tablecloth and the scarf in the first row are missing when the scaling factor is larger. Similarly, the stripes on the legs of the zebra are aliased with the increase of the scaling factor.
	}

	\subsubsection{Investigation on HEB}
	
	In this section, we mainly investigate the performance of different components in HSRNet. In Section~\ref{sec:heb}, we argue that HEB obtains a hierarchical exploration to consider the local image prior. Figure~\ref{fig:vis-heb} illustrates the processed features in different scales. To make the feature maps clear and focus on the effectiveness of network structure, we draw the pictures by the initial random network weights without further training. Figure~\ref{fig:vis-heb} (a) is the input image. Visualization images from (b) to (e) are the processed feature maps with different receptive fields. We can find the scale of features is larger and larger with the increase of receptive field. The structural textures is gradually addressed, and the high frequency information is dilated to the adjacent patches. For example, the top area of the building in (b) is flat. From (c) to (e), the lines are obvious in the feature map, and dilate to a larger scale. This diffusion phenomenon is in accordance with our local image prior hypothesis, where the refined adjacent patches can provide a good guidance for restoration. To further investigate the effectiveness of local image prior, we visualize the explored features with the guidance of hierarchical information. (b), (f)-(h) in Figure~\ref{fig:vis-heb} shows the example visualized feature maps of $F'^1$ to $F'^4$ in Eq.(\ref{eq:13}) separately. We can find with the increase of scaling factors, the aliasing is gradually suppressed, and the structural information is recovered.
	
	To further investigate the effectiveness of HEB, we build the contrast experiments with different number of the HEB. The numbers of blocks are set with $N=3, 5, 7$ and we train them under the same protocol. Table~\ref{tab:abl-heb} shows the PSNR/SSIM results with different block numbers. The performance is boosted with the increase of $N$. 
	
	\begin{table}[t]
		\centering
		\caption{Investigation on different block numbers on PSNR/SSIM with \textbf{BI} $\times4$ degradation.}
		\label{tab:abl-heb}
		\fontsize{6.5}{8}\selectfont
		\begin{tabular}{|c|c|c|c|c|}
			\hline
			\textbf{$N$}& \textbf{Set14}& \textbf{B100}	& \textbf{Urban100}	& \textbf{Manga109} \\
			\hline
			\textbf{3}	& 28.53/0.7802 & 27.54/0.7352 & 25.92/0.7810 & 30.25/0.9054 \\
			\textbf{5}	& 28.58/0.7818 & 27.58/0.7365 & 26.10/0.7868 & 30.51/0.9085 \\
			\textbf{7}	& 28.66/0.7839 & 27.63/0.7380 & 26.21/0.7904 & 30.69/0.9108 \\
			\textbf{10}	& 28.68/0.7840 & 27.64/0.7388 & 26.28/0.7934 & 30.72/0.9114 \\
			\hline
		\end{tabular}
	\end{table}
	
	\subsubsection{Investigation on MSA}
	\begin{figure}[t]
		\captionsetup[subfloat]{labelformat=empty, justification=centering}
		\begin{center}
			\subfloat[(a)]{\includegraphics[width = 0.45\linewidth]{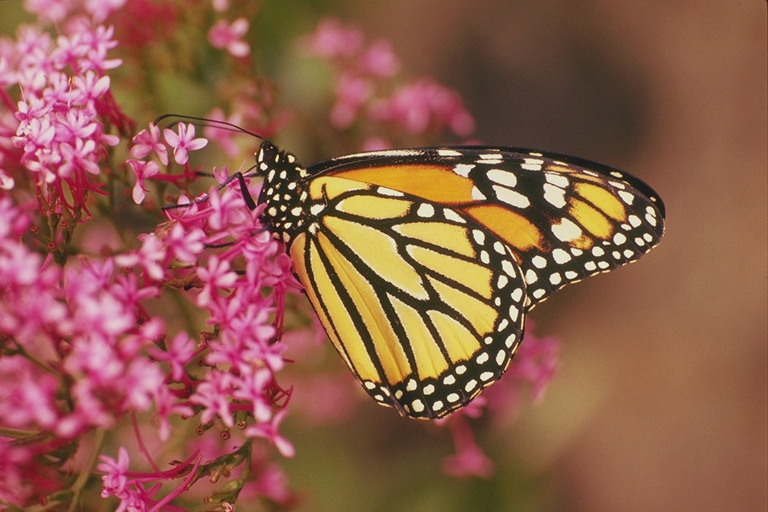}} \hspace{0.1cm}
			\subfloat[(b)]{\includegraphics[width = 0.45\linewidth]{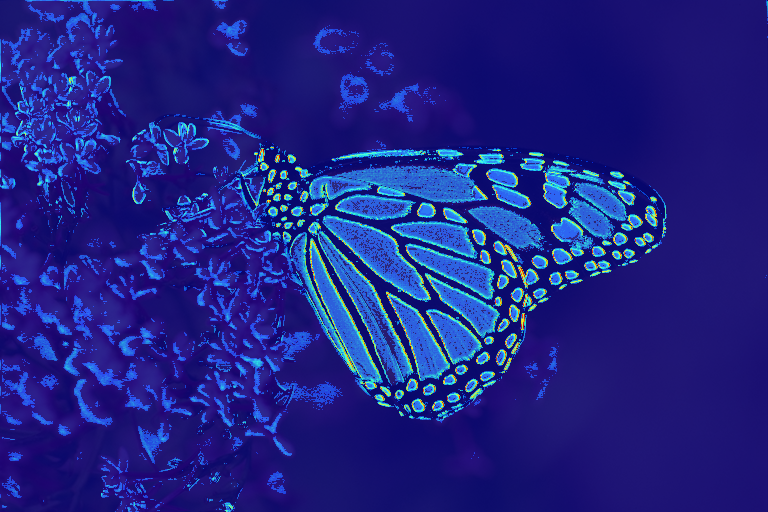}} \\ 
			\subfloat[(c)]{\includegraphics[width = 0.45\linewidth]{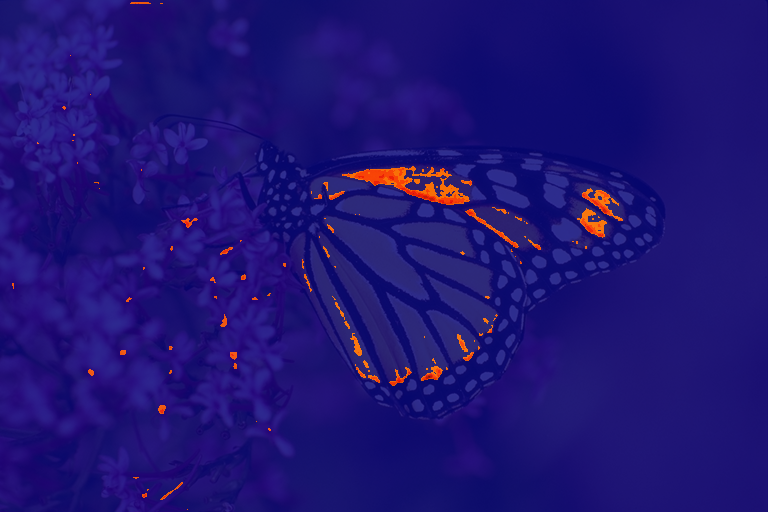}} \hspace{0.1cm}
			\subfloat[(d)]{\includegraphics[width = 0.45\linewidth]{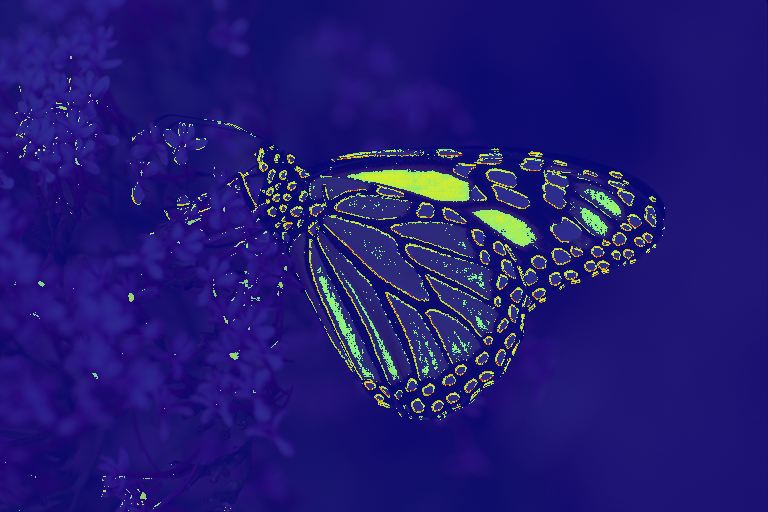}}
		\end{center}
		\caption{Visualized attention maps from different scales. (a): input image. (b)-(d): attention explored with scaling factor $\times1$, $\times2$ and $\times4$. With the increase of scaling factor, more attentions are aggregated to the high-frequency areas with structural information.}
		\label{fig:vis-msa}
	\end{figure}
	Besides the hierarchical exploration, we also investigate the effectiveness of multi-level spatial attention. In MSA, the multi-level attention aims to concentrate on the important information from different scales. Figure~\ref{fig:vis-msa} shows the visualized attention maps from different scales. (a) is the original input image. (b)-(d) are the explored attention maps with different scaling factors, which denote $F'^1_a$ to $F'^3_a$ in Eq.(\ref{eq:16}) to (\ref{eq:18}). 
	{In the figure, the receptive field becomes larger and the structural information are highly addressed with the increase of the scaling factor. We can find that in the Figure~\ref{fig:vis-msa}~(b), the attention map holds higher responses on the tiny and dense textures that captured by the small receptive field. When the scaling factor becomes larger, the structural information gets higher attention, such as the wing of the butterfly and the edges of the flowers. In this point of view, the feature map with higher scaling factor concentrates more on the structural information, and the small scaling feature focuses more on the high-frequency tiny details.}
	
	\begin{table}[t]
		\centering
		\caption{Investigation on the attention mechanism on PSNR/SSIM with \textbf{BI} $\times4$ degradation.}
		\label{tab:abl-msa}
		\fontsize{6.5}{8}\selectfont
		\begin{tabular}{|c|c|c|c|c|}
			\hline
			\textbf{MSA}& \textbf{Set14}& \textbf{B100}	& \textbf{Urban100}	& \textbf{Manga109} \\
			\hline
			\textbf{w/o}	& 28.64/0.7827 & 27.60/0.7371 & 26.14/0.7886 & 30.61/0.9091 \\
			\textbf{w}		& 28.68/0.7840 & 27.64/0.7388 & 26.28/0.7934 & 30.72/0.9114 \\
			\hline
		\end{tabular}
	\end{table}
	
	Furthermore, we also investigate the quantitative performance improved by the attention mechanism. Table~\ref{tab:abl-msa} shows the PSNR/SSIM comparisons with and without the MSA. From the results, MSA provides near 0.15 dB improvement on Urban100 and 0.1 dB on Manga109. Since Urban100 and Manga109 are high resolution benchmarks with plentiful structural information, the results demonstrate the effectiveness of MSA on the high-frequency texture recovery.
	
	\begin{table}[t]
		\centering
		\caption{Investigation on the iteration times $K$ on PSNR/SSIM with \textbf{BI} $\times4$ degradation.}
		\label{tab:abl-hqs}
		\fontsize{6.5}{8}\selectfont
		\begin{tabular}{|c|c|c|c|c|}
			\hline
			\textbf{$K$}& \textbf{Set14}& \textbf{B100}	& \textbf{Urban100}	& \textbf{Manga109} \\
			\hline
			\textbf{1}	& 28.61/0.7824 & 27.58/0.7364 & 26.11/0.7874 & 30.46/0.9082 \\
			\textbf{2}	& 28.67/0.7838 & 27.62/0.7377 & 26.22/0.7909 & 30.67/0.9105 \\
			\textbf{3}	& 28.68/0.7840 & 27.64/0.7388 & 26.28/0.7934 & 30.72/0.9114 \\
			\hline
		\end{tabular}
	\end{table}
	\subsubsection{Investigation on Iterative Design}
	\begin{figure}[t]
		\captionsetup[subfloat]{labelformat=empty, justification=centering}
		\begin{center}
			\subfloat[(a)]{\includegraphics[width = 0.22\linewidth]{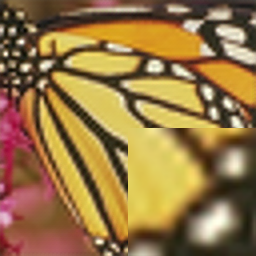}} \hspace{0.1cm}
			\subfloat[(b)]{\includegraphics[width = 0.22\linewidth]{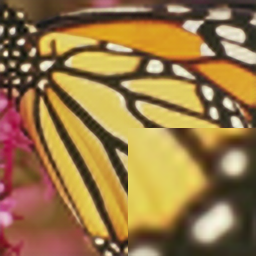}} \hspace{0.1cm}
			\subfloat[(c)]{\includegraphics[width = 0.22\linewidth]{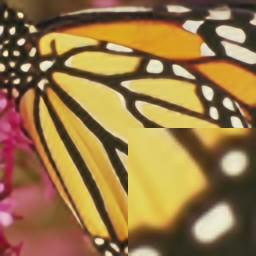}} \hspace{0.1cm}
			\subfloat[(d)]{\includegraphics[width = 0.22\linewidth]{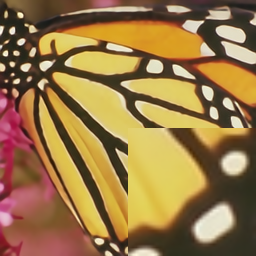}}
		\end{center}
		\caption{Restored images from different iteration times $K$. From (a) to (d), the iteration time $K$ is from $0$ to $3$.}
		\label{fig:vis-hqs}
	\end{figure}
	As discussed in the network design, we utilize HQS to analyze the image SR in the optimization perspective, and propose an iterative design to find the solution. To investigate the effectiveness of iterative design, we make the contrast comparisons with different iteration times. Table~\ref{tab:abl-hqs} shows the PSNR/SSIM results with iteration times $K=1, 2, 3$. In the table, the performance becomes better when $K$ is higher. There is near 0.05 dB PSNR improvement on Manga109 and 0.06 dB improvement on Urban100 when $K=3$.
	
	To further investigate the mechanism of iteration pattern, we compare the restored images from different iteration times. Figure~\ref{fig:vis-hqs} shows the images with $K=1, 2, 3$. We can find that with the increase of $K$, the details are richer, and the edges on the wing becomes sharper and clearer. This is in accordance with the iteration optimization.
	
	\begin{table*}[!ht]
		\centering
		\caption{Average PSNR/SSIM with degradation model \textbf{BI} $\times2$, $\times3$, and $\times4$ on five benchmarks. The best and second performances are shown in \textbf{bold} and \underline{underline}.}
		\label{tab:BI-result}
		\begin{tabular}{|c|c|c|c|c|c|c|}
			\hline
			\multirow{2}{*}{Scale}& \multirow{2}{*}{Model}&  
			Set5~\cite{set5}& Set14~\cite{set14}& B100~\cite{b100}& Urban100~\cite{urban100}& Manga109~\cite{manga109} \\
			& & PSNR/SSIM & PSNR/SSIM & PSNR/SSIM & PSNR/SSIM & PSNR/SSIM\\
			\hline
			\multirow{18}{*}{$\times2$} &SRCNN~\cite{srcnn_pami2016}&
			36.66 / 0.9542& 32.42 / 0.9063& 31.36 / 0.8879& 29.50 / 0.8946& 35.74 / 0.9661\\
			
			& FSRCNN~\cite{fsrcnn_eccv2016}&
			37.00 / 0.9558& 32.63 / 0.9088& 31.53 / 0.8920& 29.88 / 0.9020& 36.67 / 0.9694\\
			
			& VDSR~\cite{vdsr_cvpr2016}&
			37.53 / 0.9587& 33.03 / 0.9124& 31.90 / 0.8960& 30.76 / 0.9140& 37.22 / 0.9729\\
			
			& DRCN~\cite{drcn_cvpr2016}&
			37.63 / 0.9588& 33.04 / 0.9118& 31.85 / 0.8942& 30.75 / 0.9133& 37.63 / 0.9723\\
			
			&LapSRN~\cite{lapsrn_pami2019}&
			37.52 / 0.9590& 33.08 / 0.9130& 31.80 / 0.8950& 30.41 / 0.9100& 37.27 / 0.9740\\
			
			&DRRN~\cite{drrn_cvpr2017}&
			37.74 / 0.9591& 33.23 / 0.9136& 32.05 / 0.8973& 31.23 / 0.9188& 37.92 / 0.9760\\
			
			&MemNet~\cite{memnet_iccv2017}&
			37.78 / 0.9597& 33.28 / 0.9142& 32.08 / 0.8978& 31.31 / 0.9195& 37.72 / 0.9740 \\
			
			&DEGREE~\cite{degree_tip2017}&
			37.58 / 0.9587& 33.06 / 0.9123& 31.80 / 0.8974& - & - \\
			
			&CARN~\cite{carn_eccv2018}&
			37.76 / 0.9590& 33.52 / 0.9166& 32.09 / 0.8978& 31.92 / 0.9256& 38.36 / 0.9765\\
			
			&IMDN~\cite{imdn_mm2019}&
			{38.00} / 0.9605& {33.63} / {0.9177}& {32.19 / 0.8996}& {32.17 / 0.9283}& \underline{38.88 / 0.9774}\\
			
			&RAN~\cite{ran_csvt2019}&
			37.58 / 0.9592 &33.10 / 0.9133 &31.92 / 0.8963& -& -\\
			
			&DNCL~\cite{dncl_csvt2019}&
			37.65 / 0.9599 &33.18 / 0.9141 &31.97 / 0.8971 &30.89 / 0.9158& - \\
			
			&FilterNet~\cite{filternet_csvt2020}&
			37.86 / 0.9610 &33.34 / 0.9150 &32.09 / 0.8990 &31.24 / 0.9200& - \\
			
			&MRFN~\cite{mrfn_tmm2020}&
			37.98 / 0.9611 &33.41 / 0.9159 &32.14 / 0.8997 &31.45 / 0.9221 &38.29 / 0.9759\\
			
			&SeaNet-baseline~\cite{seanet_tip2020}&
			37.99 / {0.9607} & 33.60 / 0.9174 & 32.18 / 0.8995 & 32.08 / 0.9276 & 38.48 / 0.9768 \\
			
			&MSRN~\cite{msrn_eccv2018}&
			\textbf{38.08 / 0.9607} & \underline{33.70 / 0.9186} & \underline{32.23 / 0.9002} & \underline{32.29 / 0.9303} & 38.69 / 0.9772 \\
			
			&HSRNet&
			\underline{38.07 / 0.9607} & \textbf{33.78 / 0.9197} & \textbf{32.26 / 0.9006} & \textbf{32.53 / 0.9320} & \textbf{38.90 / 0.9776} \\
			
			\hline
			
			\multirow{17}{*}{$\times3$}& SRCNN~\cite{srcnn_pami2016} &
			32.75 / 0.9090& 29.28 / 0.8209& 28.41 / 0.7863& 26.24 / 0.7989& 30.59 / 0.9107\\
			
			&FSRCNN~\cite{fsrcnn_eccv2016}&
			33.16 / 0.9140& 29.43 / 0.8242& 28.53 / 0.7910& 26.43 / 0.8080& 30.98 / 0.9212\\
			
			&VDSR~\cite{vdsr_cvpr2016}&
			33.66 / 0.9213& 29.77 / 0.8314& 28.82 / 0.7976& 27.14 / 0.8279& 32.01 / 0.9310\\
			
			&DRCN~\cite{drcn_cvpr2016}&
			33.82 / 0.9226& 29.76 / 0.8311& 28.80 / 0.7963& 27.15 / 0.8276& 32.31 / 0.9328\\
			
			&DRRN~\cite{drrn_cvpr2017}&
			34.03 / 0.9244& 29.96 / 0.8349& 28.95 / 0.8004& 27.53 / 0.8378& 32.74 / 0.9390\\
			
			&MemNet~\cite{memnet_iccv2017}&
			34.09 / 0.9248& 30.00 / 0.8350& 28.96 / 0.8001& 27.56 / 0.8376& 32.51 / 0.9369 \\
			
			&DEGREE~\cite{degree_tip2017}&
			33.76 / 0.9211& 29.82 / 0.8326& 28.74 / 0.7950& - & - \\
			
			&CARN~\cite{carn_eccv2018}&
			34.29 / 0.9255& 30.29 / 0.8407& 29.06 / 0.8034& 28.06 / 0.8493& 33.50 / 0.9440\\
			
			&IMDN~\cite{imdn_mm2019}&
			34.36 / 0.9270& 30.32 / 0.8417& {29.09 / 0.8046}& 28.17 / 0.8519& {33.61 / 0.9445}\\
			
			&RAN~\cite{ran_csvt2019}&
			33.71 / 0.9223 & 29.84 / 0.8326 & 28.84 / 0.7981 & - & -\\
			
			&DNCL~\cite{dncl_csvt2019}&
			33.95 / 0.9232 & 29.93 / 0.8340 & 28.91 / 0.7995 & 27.27 / 0.8326 & -\\
			
			&FilterNet~\cite{filternet_csvt2020}&
			34.08 / 0.9250 & 30.03 / 0.8370 & 28.95 / 0.8030 & 27.55 / 0.8380 & -\\
			
			&MRFN~\cite{mrfn_tmm2020}&
			34.21 / 0.9267 & 30.03 / 0.8363 & 28.99 / 0.8029 & 27.53 / 0.8389 & 32.82 / 0.9396 \\
			
			&SeaNet-baseline~\cite{seanet_tip2020}&
			{34.36 / 0.9280} & {30.34 / 0.8428} & {29.09 / 0.8053} & 28.17 / 0.8527 & 33.40 / 0.9444 \\
			
			&MSRN~\cite{msrn_eccv2018}&
			\underline{34.46 / 0.9278} & \textbf{30.41 / 0.8437} & \underline{29.15 / 0.8064} & \underline{28.33 / 0.8561} & \underline{33.67 / 0.9456} \\
			
			&HSRNet&
			\textbf{34.47 / 0.9278} & \underline{30.40 / 0.8435} & \textbf{29.15 / 0.8066} & \textbf{28.42 / 0.8579} & \textbf{33.75 / 0.9459} \\
			
			\hline
			\multirow{19}{*}{$\times4$}&SRCNN~\cite{srcnn_pami2016}&
			30.48 / 0.8628& 27.49 / 0.7503& 26.90 / 0.7101& 24.52 / 0.7221& 27.66 / 0.8505\\
			
			&FSRCNN~\cite{fsrcnn_eccv2016}&
			30.71 / 0.8657& 27.59 / 0.7535& 26.98 / 0.7150& 24.62 / 0.7280& 27.90 / 0.8517\\
			
			&VDSR~\cite{vdsr_cvpr2016}&
			31.35 / 0.8838& 28.01 / 0.7674& 27.29 / 0.7251& 25.18 / 0.7524& 28.83 / 0.8809\\
			
			&DRCN~\cite{drcn_cvpr2016}&
			31.53 / 0.8854& 28.02 / 0.7670& 27.23 / 0.7233& 25.14 / 0.7510& 28.98 / 0.8816\\
			
			&LapSRN~\cite{lapsrn_pami2019}&
			31.54 / 0.8850& 28.19 / 0.7720& 27.32 / 0.7280& 25.21 / 0.7560& 29.09 / 0.8845\\
			
			&DRRN~\cite{drrn_cvpr2017}&
			31.68 / 0.8888& 28.21 / 0.7720& 27.38 / 0.7284& 25.44 / 0.7638& 29.46 / 0.8960\\
			
			&MemNet~\cite{memnet_iccv2017}&
			31.74 / 0.8893& 28.26 / 0.7723& 27.40 / 0.7281& 25.50 / 0.7630& 29.42 / 0.8942 \\
			
			&SRDenseNet~\cite{srdensenet_iccv2017}&
			32.02 / 0.8934& 28.50 / 0.7782& 27.53 / 0.7337& 26.05 / 0.7819& - \\
			
			&DEGREE~\cite{degree_tip2017}&
			31.47 / 0.8837& 28.10 / 0.7669& 27.20 / 0.7216& - & - \\
			
			&CARN~\cite{carn_eccv2018}&
			32.13 / 0.8937& {28.60} / 0.7806& 27.58 / 0.7349& {26.07} / 0.7837& {30.47} / {0.9084}\\
			
			&IMDN~\cite{imdn_mm2019}&
			{32.21 / 0.8948}& 28.58 / {0.7811}& 27.56 / 0.7353& 26.04 / {0.7838}& {30.45} / 0.9075\\
			
			&RAN~\cite{ran_csvt2019}&
			31.43 / 0.8847 & 28.09 / 0.7691 & 27.31 / 0.7260 & - & -\\
			
			&DNCL~\cite{dncl_csvt2019}&
			31.66 / 0.8871 & 28.23 / 0.7717 & 27.39 / 0.7282 & 25.36 / 0.7606 & -\\
			
			&FilterNet~\cite{filternet_csvt2020}&
			31.74 / 0.8900 & 28.27 / 0.7730 & 27.39 / 0.7290 & 25.53 / 0.7680 & -\\
			
			&MRFN~\cite{mrfn_tmm2020}&
			31.90 / 0.8916 & 28.31 / 0.7746 & 27.43 / 0.7309 & 25.46 / 0.7654 & 29.57 / 0.8962 \\
			
			&SeaNet-baseline~\cite{seanet_tip2020}&
			32.18 / 0.8948 & {28.61 / 0.7822} & {27.57 / 0.7359} & 26.05 / 0.7896 & 30.44 / {0.9088} \\
			
			&MSRN~\cite{msrn_eccv2018}&
			\underline{32.26 / 0.8960} & \underline{28.63 / 0.7836} & \underline{27.61 / 0.7380} & \underline{26.22 / 0.7911} & \underline{30.57 / 0.9103} \\
			
			&HSRNet&
			\textbf{32.28 / 0.8960} & \textbf{28.68 / 0.7840} & \textbf{27.64 / 0.7388} & \textbf{26.28 / 0.7934} & \textbf{30.72 / 0.9114} \\
			
			\hline
		\end{tabular}
	\end{table*}
	
	\subsection{Results}
	We mainly compare HSRNet with advanced networks to show the performance: SRCNN~\cite{srcnn_pami2016}, FSRCNN~\cite{fsrcnn_eccv2016}, VDSR~\cite{vdsr_cvpr2016}, DRCN~\cite{drcn_cvpr2016}, DRRN~\cite{drrn_cvpr2017}, LapSRN~\cite{lapsrn_pami2019}, MemNet~\cite{memnet_iccv2017}, DEGREE~\cite{degree_tip2017}, CARN~\cite{carn_eccv2018}, IMDN~\cite{imdn_mm2019}, RAN~\cite{ran_csvt2019}, DNCL~\cite{dncl_csvt2019}, FilterNet~\cite{filternet_csvt2020}, MRFN~\cite{mrfn_tmm2020}, SeaNet~\cite{seanet_tip2020}, and MSRN~\cite{msrn_eccv2018}. Table~\ref{tab:BI-result} shows the average PSNR/SSIM with degradation \textbf{BI} $\times2$, $\times3$ and $\times4$ on five testing benchmarks.
	
	In the table, we can find that our HSRNet achieves almost best performance on all benchmarks. Compared with MSRN, our network achieves near 0.3 dB improvement on Urban100 and Manga109 with \textbf{BI} $\times2$ degradation. When the scaling factor is larger, our network achieves near 0.1 and 0.2 dB improvement on Manga109. It is worth noting that our network achieves significant improvements on Urban100 and Manga109 with all scaling factors. Since Urban100 and Manga109 are composed of building and comic instances with plentiful repetitive edges and lines, the higher PSNR/SSIM shows our network's superiority on structural information recovery.
	
	\begin{table}[t]
		\centering
		\caption{Average PSNR/SSIM, parameters and MACs comparisons with hierarchical networks with \textbf{BI} $\times4$ degradation.}
		\label{tab:multiscale}
		\fontsize{6.5}{8}\selectfont
		\begin{tabular}{|c|c|c|c|c|}
			\hline
			\textbf{Method}& \textbf{LapSRN~\cite{lapsrn_pami2019}}& \textbf{MSRN~\cite{msrn_eccv2018}}	& \textbf{LWSR~\cite{s_LWSR_tip2020}}	& \textbf{HSRNet} \\
			\hline
			\hline
			\textbf{Param(M)}& 0.813& 6.373& 2.277& 1.285 \\
			\textbf{MACs(G)}& 149.4& 368.6& 131.1& 203.2  \\
			\hline
			\hline
			\textbf{Set5}&		31.54 / 0.8850& 32.26 / 0.8960& 32.28 / 0.8960& 32.28 / 0.8960\\
			\textbf{Set14}&		28.19 / 0.7720& 28.63 / 0.7836& 28.34 / 0.7800& 28.68 / 0.7840\\
			\textbf{B100}& 		27.32 / 0.7280& 27.61 / 0.7380& 27.61 / 0.7380& 27.64 / 0.7388\\
			\textbf{Urban100}& 	25.21 / 0.7560& 26.22 / 0.7911& 26.19 / 0.8910& 26.28 / 0.7934\\
			\hline
		\end{tabular}
	\end{table}
	
	Specially, we mainly compare HSRNet with other hierarchical networks to demonstrate the effectiveness of our design. Table~\ref{tab:multiscale} shows the average PSNR/SSIM, parameters and computation complexity with different hierarchical networks: LapSRN~\cite{lapsrn_pami2019}, MSRN~\cite{msrn_eccv2018}, and LWSR~\cite{s_LWSR_tip2020}. Herein, we calculate the computation complexity by the number of multiply-accumulate operations~(MACs). We model the complexities by restoring a $720$P ($1280\times720$) image with scaling factor $\times4$. The MACs is calculated by PyTorch-OpCounter~\footnote{https://github.com/Lyken17/pytorch-OpCounter}. In Table~\ref{tab:multiscale}, our network requires near 55\% MACs and 20\% parameters than MSRN, and achieves better restoration performance. Compared with LWSR, our network holds near half parameters and achieves 0.3 dB higher PSNR on Set14. In this point of view, our hierarchical exploration proves to be an efficient design.
	
	\begin{table}[t]
		\centering
		\caption{Average PSNR/SSIM, parameters and MACs comparisons with iterative networks with \textbf{BI} $\times2$ degradation.}
		\label{tab:feedback}
		\fontsize{6.5}{8}\selectfont
		\begin{tabular}{|c|c|c|c|c|}
			\hline
			\textbf{Method}& \textbf{D-DBPN~\cite{dbpn_pami2020}}& \textbf{SRFBN~\cite{srfbn_cvpr2019}}	& \textbf{USRNet~\cite{usrnet_cvpr2020}}	& \textbf{HSRNet} \\
			\hline
			\hline
			\textbf{Param(M)}   & 5.95& 2.14& 17.01&  1.26\\
			\textbf{MACs(G)}    & 3746.2& 5043.5& 8545.8&  808.2 \\
			\hline
			\hline
			\textbf{Set5}&		38.09 / 0.9600& 38.11 / 0.9609& 37.76/0.9599& 38.07 / 0.9607\\
			\textbf{Set14}&		33.85 / 0.9190& 33.82 / 0.9196& 33.43/0.9159& 33.78 / 0.9197\\
			\textbf{B100}& 		32.27 / 0.9000& 32.29 / 0.9010& 32.09/0.8985& 32.26 / 0.9006\\
			\textbf{Urban100}& 	32.55 / 0.9324& 32.62 / 0.9328& 31.78/0.9259& 32.53 / 0.9320\\
			\hline
		\end{tabular}
	\end{table}
	
	\begin{figure*}[t]
		\captionsetup[subfloat]{labelformat=empty, justification=centering}
		\begin{center}
			\subfloat[HR\linebreak{(PSNR/SSIM)}]{\includegraphics[width = 0.15\linewidth]{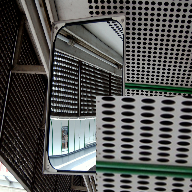}} \hspace{0.01cm}
			\subfloat[LR\linebreak{(19.23/0.6343)}]{\includegraphics[width = 0.15\linewidth]{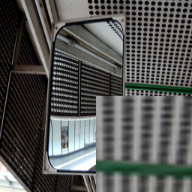}} \hspace{0.01cm}
			\subfloat[LapSRN\linebreak{(21.11/0.8063)}]{\includegraphics[width = 0.15\linewidth]{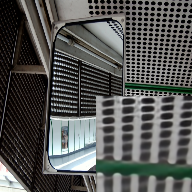}} \hspace{0.01cm}
			\subfloat[VDSR\linebreak{(21.12/0.7968)}]{\includegraphics[width = 0.15\linewidth]{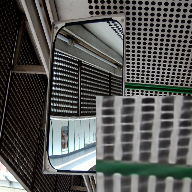}} \hspace{0.01cm}
			\subfloat[MSRN\linebreak{(22.50/0.8478)}]{\includegraphics[width = 0.15\linewidth]{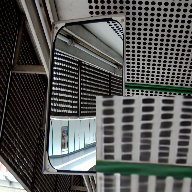}} \hspace{0.01cm}
			\subfloat[HSRNet\linebreak{(\textbf{22.83/0.8538})}]{\includegraphics[width = 0.15\linewidth]{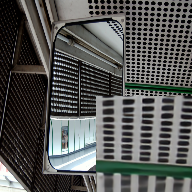}}
			\\ \vspace{-0.3cm}
			\subfloat[HR\linebreak{(PSNR/SSIM)}]{\includegraphics[width = 0.15\linewidth]{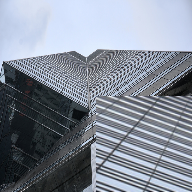}} \hspace{0.01cm}
			\subfloat[LR\linebreak{(19.03/0.6311)}]{\includegraphics[width = 0.15\linewidth]{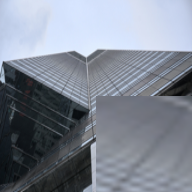}} \hspace{0.01cm}
			\subfloat[LapSRN\linebreak{(19.93/0.7175)}]{\includegraphics[width = 0.15\linewidth]{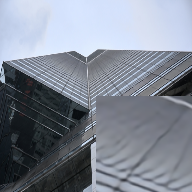}} \hspace{0.01cm}
			\subfloat[VDSR\linebreak{(19.94/0.7189)}]{\includegraphics[width = 0.15\linewidth]{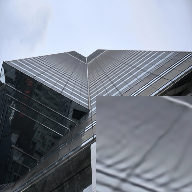}} \hspace{0.01cm}
			\subfloat[MSRN\linebreak{(21.12/0.7719)}]{\includegraphics[width = 0.15\linewidth]{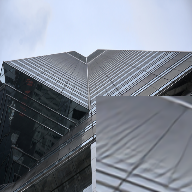}} \hspace{0.01cm}
			\subfloat[HSRNet\linebreak{(\textbf{21.40/0.7785})}]{\includegraphics[width = 0.15\linewidth]{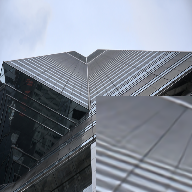}} \hspace{0.01cm}
			\\ \vspace{-0.3cm}
			\subfloat[HR\linebreak{(PSNR/SSIM)}]{\includegraphics[width = 0.15\linewidth]{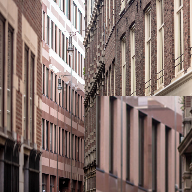}} \hspace{0.01cm}
			\subfloat[LR\linebreak{(23.07/0.6773)}]{\includegraphics[width = 0.15\linewidth]{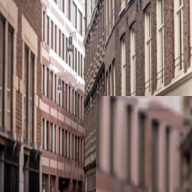}} \hspace{0.01cm}
			\subfloat[LapSRN\linebreak{(24.56/0.7669)}]{\includegraphics[width = 0.15\linewidth]{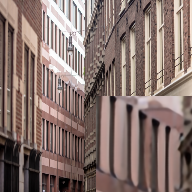}} \hspace{0.01cm}
			\subfloat[VDSR\linebreak{(24.42/0.7620)}]{\includegraphics[width = 0.15\linewidth]{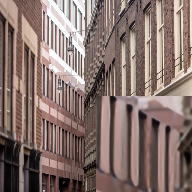}} \hspace{0.01cm}
			\subfloat[MSRN\linebreak{(24.98/0.7857)}]{\includegraphics[width = 0.15\linewidth]{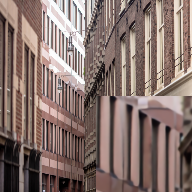}} \hspace{0.01cm}
			\subfloat[HSRNet\linebreak{(\textbf{25.32/0.7947})}]{\includegraphics[width = 0.15\linewidth]{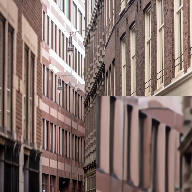}} \hspace{0.01cm}
			\\ \vspace{-0.3cm}
			\subfloat[HR\linebreak{(PSNR/SSIM)}]{\includegraphics[width = 0.15\linewidth]{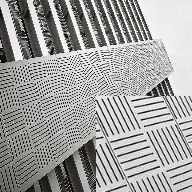}} \hspace{0.01cm}
			\subfloat[LR\linebreak{(15.11/0.4153)}]{\includegraphics[width = 0.15\linewidth]{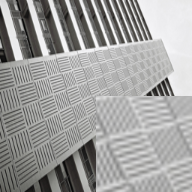}} \hspace{0.01cm}
			\subfloat[LapSRN\linebreak{(16.93/0.6034)}]{\includegraphics[width = 0.15\linewidth]{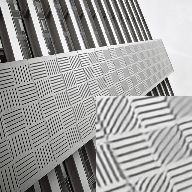}} \hspace{0.01cm}
			\subfloat[VDSR\linebreak{(16.87/0.5988)}]{\includegraphics[width = 0.15\linewidth]{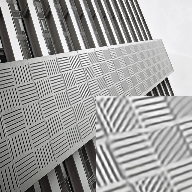}} \hspace{0.01cm}
			\subfloat[MSRN\linebreak{(17.64/0.6540)}]{\includegraphics[width = 0.15\linewidth]{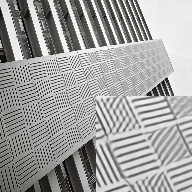}} \hspace{0.01cm}
			\subfloat[HSRNet\linebreak{(\textbf{17.87/0.6651})}]{\includegraphics[width = 0.15\linewidth]{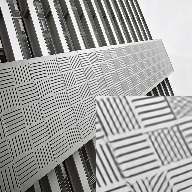}} \hspace{0.01cm}
		\end{center}
		\caption{Visualization comparisons on Urban100 with \textbf{BI} $\times4$.}
		\label{fig:vis-urban100}
	\end{figure*}
	
	Furthermore, we also compare HSRNet with other iterative networks to show the effectiveness of our HQS pattern. Table~\ref{tab:feedback} shows the performance comparisons with D-DBPN~\cite{dbpn_pami2020}, SRFBN~\cite{srfbn_cvpr2019}, and USRNet~\cite{usrnet_cvpr2020}. From the results, HSRNet achieves competitive performance than D-DBPN and SRFBN. It is worth noting that our network only holds 21\% parameters and MACs than D-DBPN, and 58\% parameters and 16\% MACs than SRFBN. This is because D-DBPN and SRFBN explicitly increase the image resolution while HSRNet performs the optimization in the specific space, which significantly decrease the computation cost. USRNet is another iterative network based on the HQS optimization. Compared with USRNet, our network achieves near 0.8 dB PSNR higher on Urban100, 0.3 dB higher on Set5 and Set14, with only 7\% parameters and 9\% MACs. In this point of view, our HSRNet enjoys a more efficient network design to perform the optimization.
	
	We also perform the visualization comparison to show the effectiveness of structural information restoration and aliasing suppression. Figure~\ref{fig:vis-urban100} shows the example aliasing images in Urban100. We can find the aliasing is heavy in LR instances. Compared with other methods, our network can suppress the aliasing and restore the accurate textures more effectively. This is because the buildings contain abundant repeated information, and the adjacent patches usually holds similar textures. In this point of view, local image prior can provide a good guidance for image restoration. For the first row in the figure, the textures of the holes are similar, and our network can distinguish more accurate circles under the aliasing. For the last row in the figure, the lines are similar, and we can recover more correct lines.
	
	\begin{figure*}[t]
		\captionsetup[subfloat]{labelformat=empty, justification=centering}
		\begin{center}
			\subfloat[HR\linebreak{(PSNR/SSIM)}]{							\includegraphics[width = 0.151\linewidth]{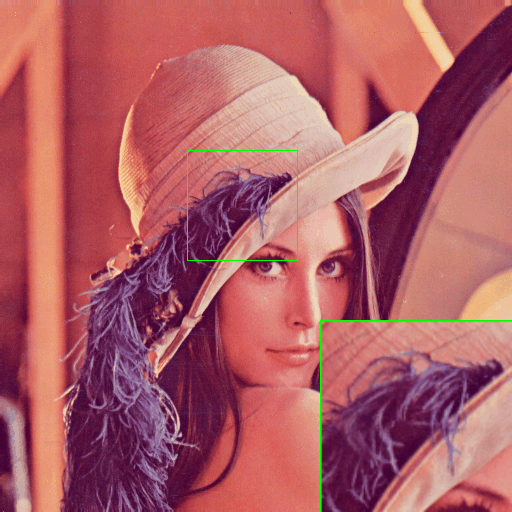}}
			\subfloat[LR\linebreak{(28.50/0.7866)}]{						\includegraphics[width = 0.151\linewidth]{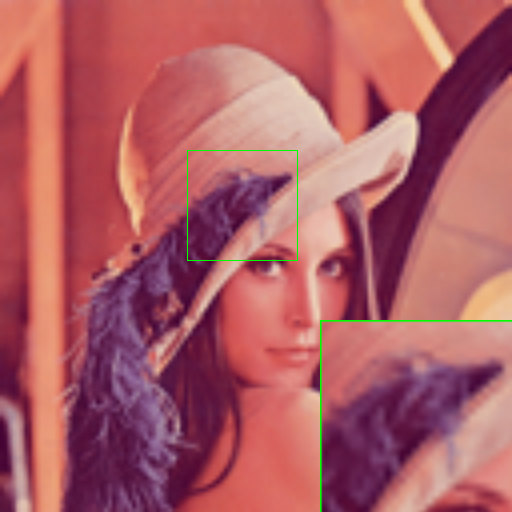}}
			\subfloat[VDSR~\cite{vdsr_cvpr2016}\linebreak{(30.22/0.8242)}]{	\includegraphics[width = 0.151\linewidth]{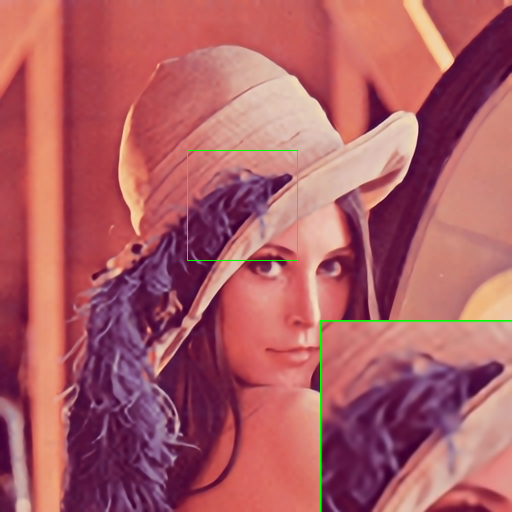}}
			\subfloat[CrossSRN~\cite{crosssrn}\linebreak{(30.77/0.8349)}]{	\includegraphics[width = 0.151\linewidth]{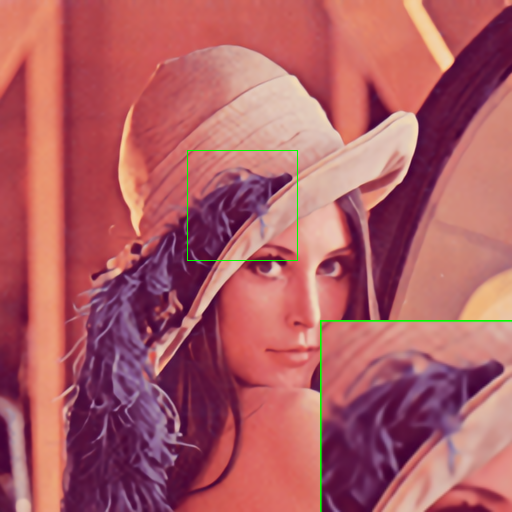}}
			\subfloat[MSRN~\cite{msrn_eccv2018}\linebreak{(30.66/0.8337)}]{	\includegraphics[width = 0.151\linewidth]{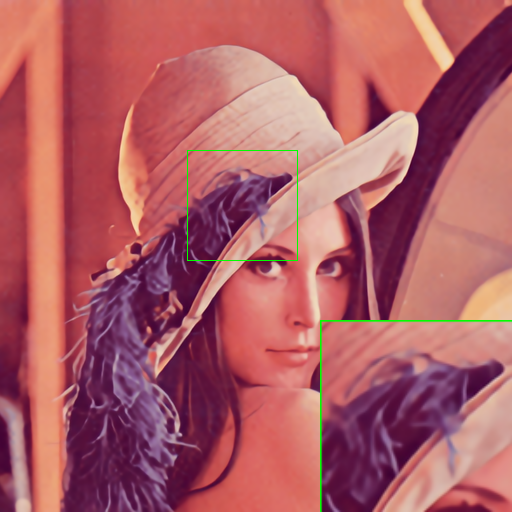}}
			\subfloat[HSRNet\linebreak{(\textbf{30.90/0.8375})}]{			\includegraphics[width = 0.151\linewidth]{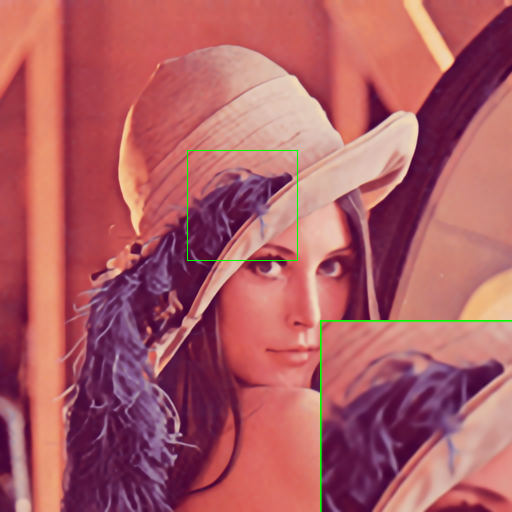}}
			\\ \vspace{-0.3cm}
			\subfloat[HR\linebreak{(PSNR/SSIM)}]{							\includegraphics[width = 0.151\linewidth]{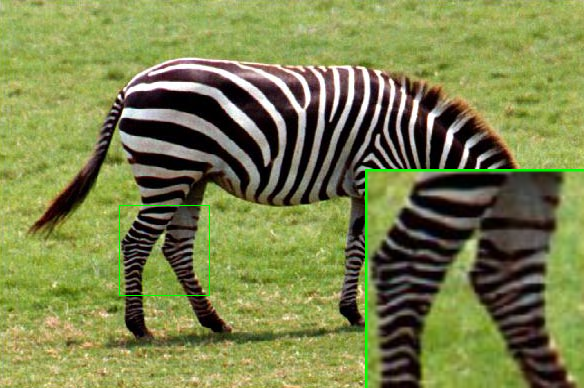}}
			\subfloat[LR\linebreak{(23.10/0.6901)}]{						\includegraphics[width = 0.151\linewidth]{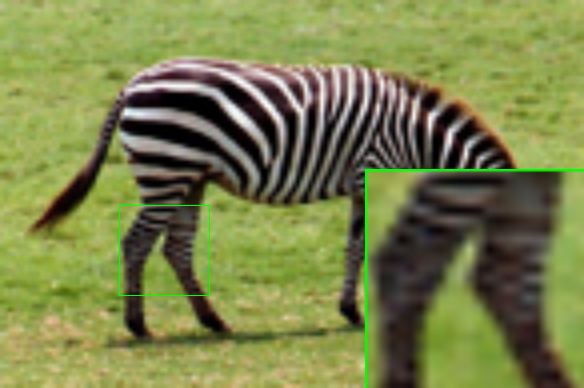}}
			\subfloat[VDSR~\cite{vdsr_cvpr2016}\linebreak{(25.47/0.7665)}]{	\includegraphics[width = 0.151\linewidth]{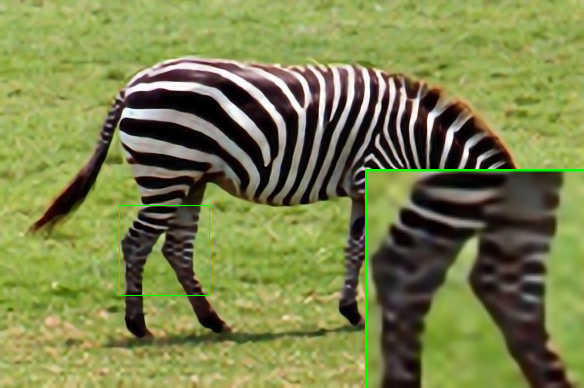}}
			\subfloat[CrossSRN~\cite{crosssrn}\linebreak{(26.30/0.7842)}]{	\includegraphics[width = 0.151\linewidth]{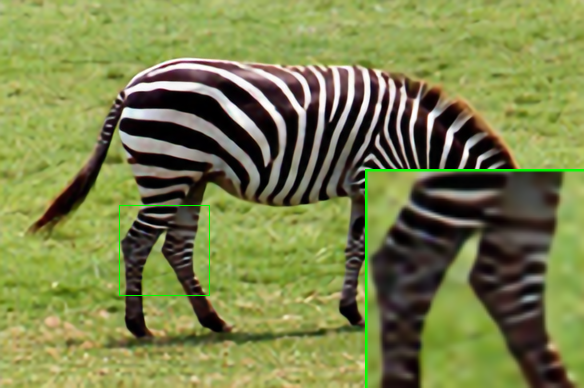}}
			\subfloat[MSRN~\cite{msrn_eccv2018}\linebreak{(26.19/0.7842)}]{	\includegraphics[width = 0.151\linewidth]{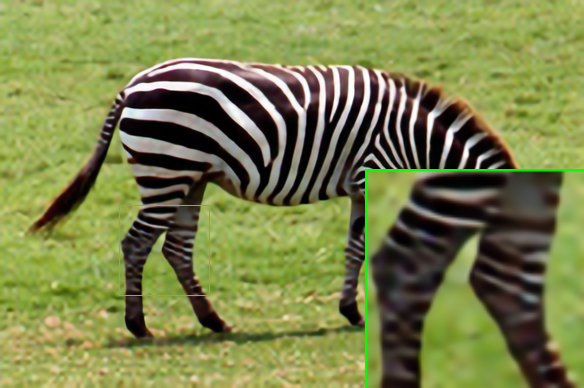}}
			\subfloat[HSRNet\linebreak{(\textbf{26.41/0.7860})}]{			\includegraphics[width = 0.151\linewidth]{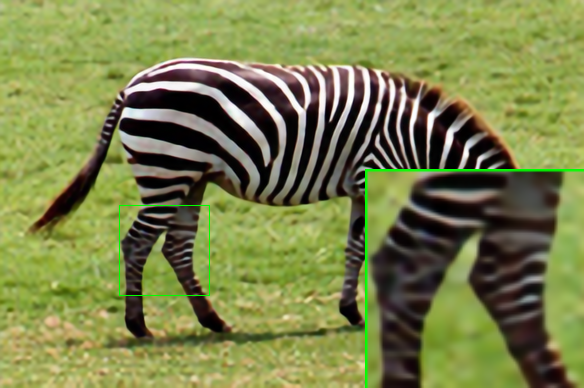}}
		\end{center}
		\caption{{Visualization comparisons on Set14 with \textbf{BI} $\times4$.}}
		\label{fig:vis-set14}
	\end{figure*}
	
	\begin{figure*}[t]
		\captionsetup[subfloat]{labelformat=empty, justification=centering}
		\begin{center}
			\subfloat[HR\linebreak{(PSNR/SSIM)}]{										\includegraphics[width = 0.151\linewidth]{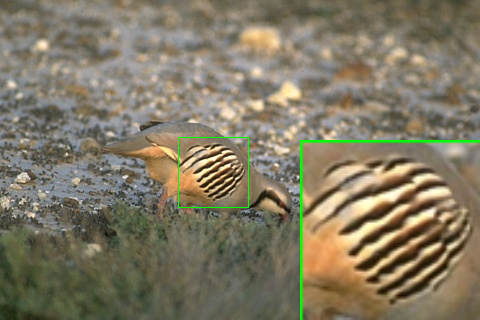}}
			\subfloat[LR\linebreak{(27.51/0.8346)}]{									\includegraphics[width = 0.151\linewidth]{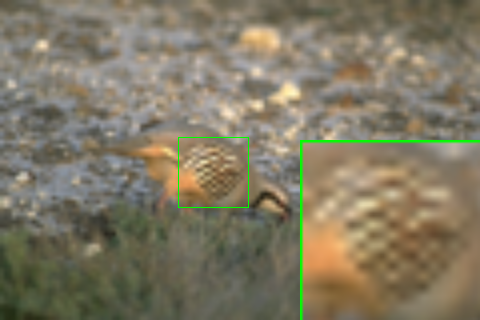}}
			\subfloat[VDSR~\cite{vdsr_cvpr2016}\linebreak{(28.37/0.8684)}]{				\includegraphics[width = 0.151\linewidth]{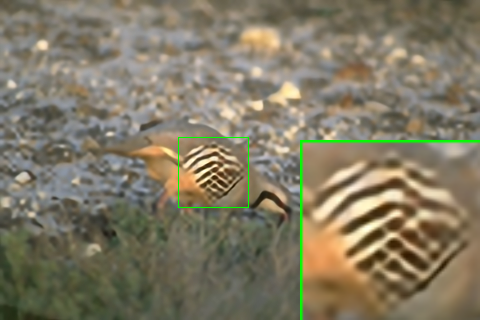}}
			\subfloat[CrossSRN~\cite{crosssrn}\linebreak{(28.50/0.8757)}]{				\includegraphics[width = 0.151\linewidth]{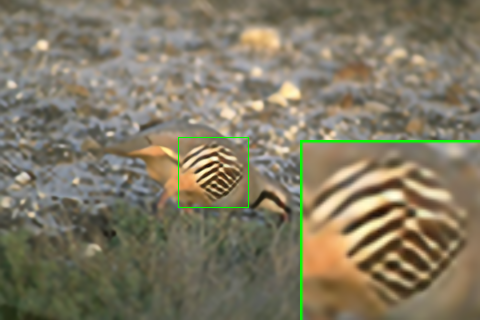}}
			\subfloat[MSRN~\cite{msrn_eccv2018}\linebreak{(29.13/0.8790)}]{			 	\includegraphics[width = 0.151\linewidth]{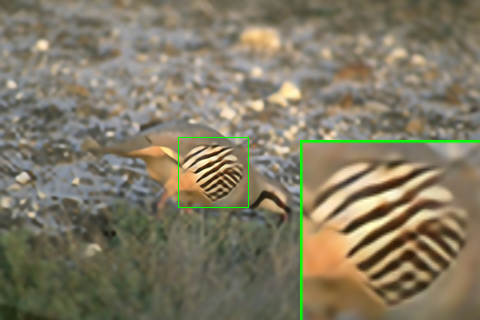}}
			\subfloat[HSRNet\linebreak{(\textbf{29.88/0.8818})}]{						\includegraphics[width = 0.151\linewidth]{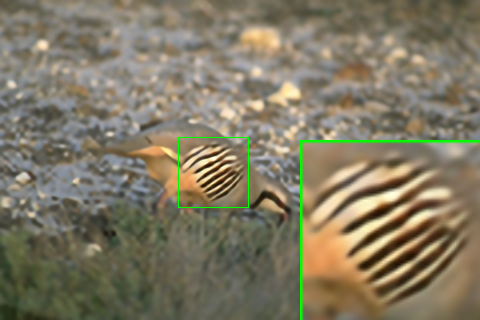}}
			\\ \vspace{-0.3cm}
			\subfloat[HR\linebreak{(PSNR/SSIM)}]{										\includegraphics[width = 0.151\linewidth]{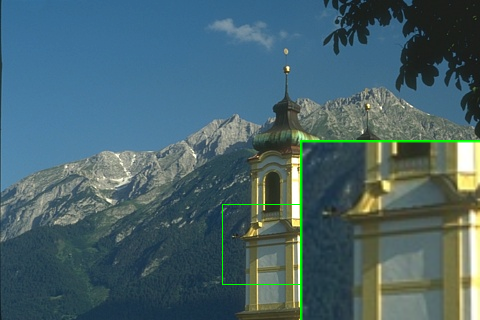}}
			\subfloat[LR\linebreak{(26.15/0.7523)}]{									\includegraphics[width = 0.151\linewidth]{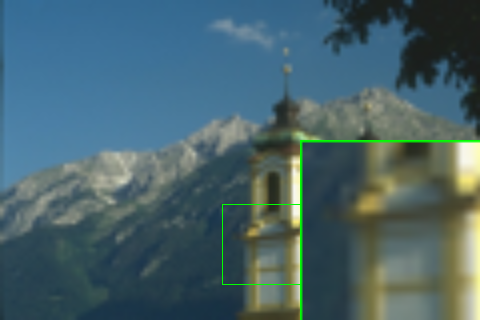}}
			\subfloat[VDSR~\cite{vdsr_cvpr2016}\linebreak{(27.73/0.8077)}]{				\includegraphics[width = 0.151\linewidth]{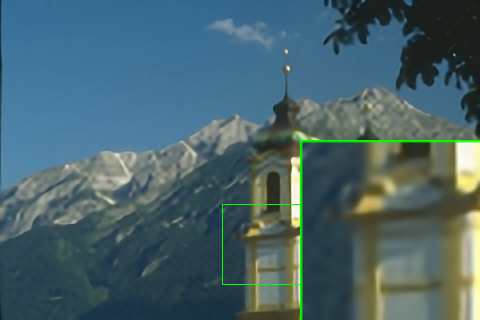}}
			\subfloat[CrossSRN~\cite{crosssrn}\linebreak{(28.23/0.8222)}]{				\includegraphics[width = 0.151\linewidth]{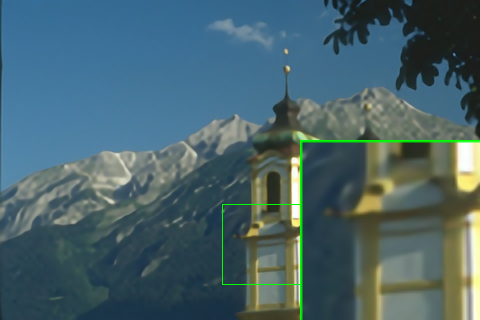}}
			\subfloat[MSRN~\cite{msrn_eccv2018}\linebreak{(28.01/0.8197)}]{			 	\includegraphics[width = 0.151\linewidth]{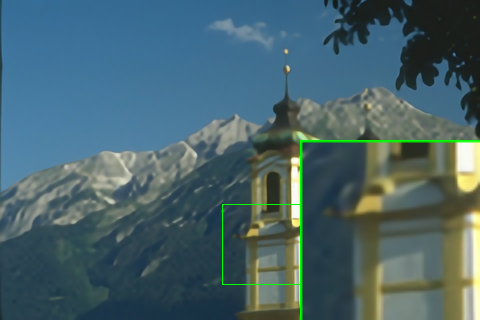}}
			\subfloat[HSRNet\linebreak{(\textbf{28.23/0.8224})}]{						\includegraphics[width = 0.151\linewidth]{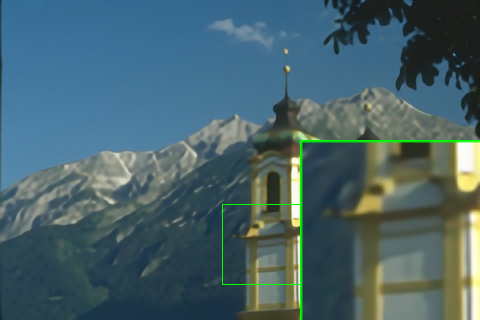}}
		\end{center}
		\caption{{Visualization comparisons on B100 with \textbf{BI} $\times4$.}}
		\label{fig:vis-b100}
	\end{figure*}
	
	{
		To demonstrate the effectiveness on different objects and scenarios, we compare our method with different methods on big and small animals, people and building from Set14~\cite{set14} and B100~\cite{b100} benchmarks, which contains the natural and the man-made scenarios. Figure~\ref{fig:vis-set14} and \ref{fig:vis-b100} shows the visualization comparisons. In Figure~\ref{fig:vis-set14}, the hairs are mixed by the bicubic degradation. HSRNet restores more accurate edges than other works. The zebra's leg suffers from aliasing effect which makes it hard to restore the correct textures. Compared with other works, the result of HSRNet is closest to the HR image. Figure~\ref{fig:vis-b100} also demonstrates the effectiveness of our method in the natural scenario. We can find that the wing of the bird is highly mixed by the degradation. HSRNet can suppress the aliasing effect and restore better textures than other works. Similarly, the lines of the building can be better reconstructed by our HSRNet. In this point of view, HSRNet is effective on different objects and scenarios.
	}

	\subsection{{Discussion}}
	\begin{figure}[t]
		\centering
		\includegraphics[width = \linewidth]{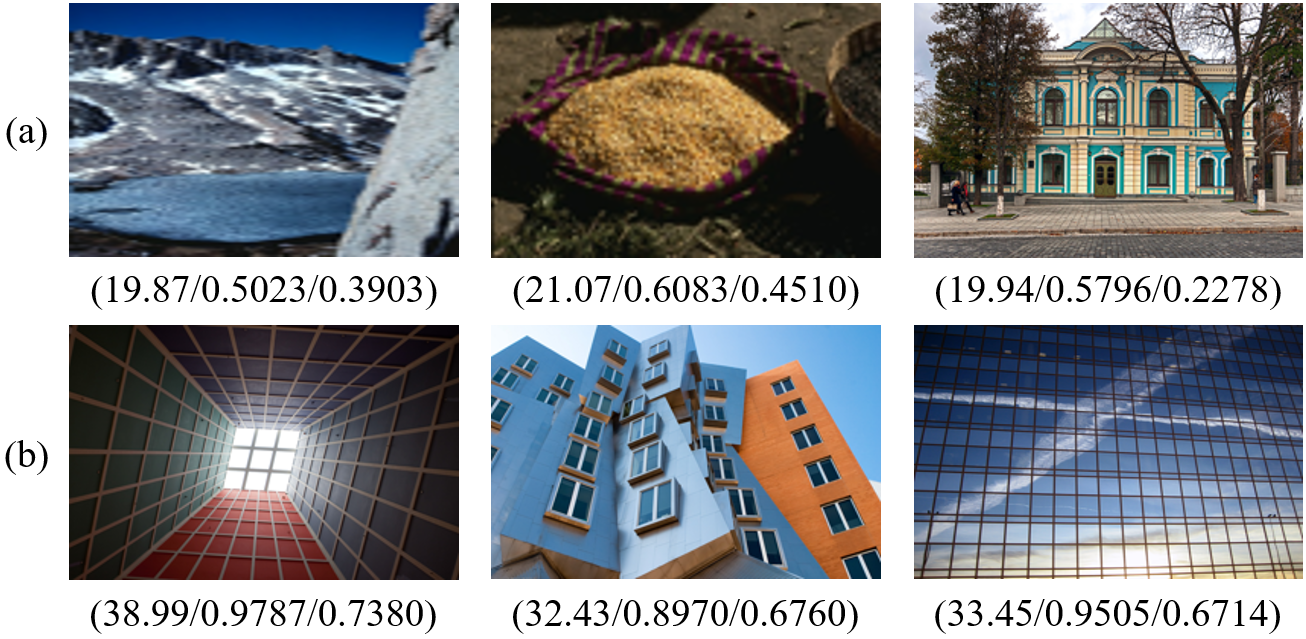}
		\caption{{Comparison between the local self-similarity and the restoration performance. The values denote the PSNR/SSIM/LSS results of the image. (a) row shows the images with lower LSS and (b) row shows the images with higher LSS.}}
		\label{fig:r2q2}
	\end{figure}

	\begin{table}[t]
		\centering
		\caption{{Investigation on the PLCC/SRCC results between PSNR and LSS on different benchmarks with $\textbf{BI}\times4$.}}
		\label{tab:r2q3}
		\fontsize{6.5}{8}\selectfont
		\begin{tabular}{|c|c|c|c|c|}
			\hline
			\textbf{Images} & \textbf{Set14}& \textbf{B100}	& \textbf{Urban100}	& \textbf{Manga109} \\
			\hline
			\textbf{LR}	    & 0.4423/0.3670& 0.4807/0.4571& 0.6124/0.5961& 0.5905/0.5433\\
			\textbf{SR}		& 0.6337/0.5824& 0.7620/0.8039& 0.7130/0.7259& 0.8084/0.7651\\
			\textbf{HR}     & 0.8553/0.7758& 0.9149/0.9326& 0.8205/0.8447& 0.8719/0.8506\\
			\hline
		\end{tabular}
	\end{table}
	
	{
		The local self-similarity (LSS) is the key point for HSRNet to restore the HR image. To model the local self-similarity, we learn from the conception of self-similarity descriptor~\cite{lss} and propose a method to estimate the similarity of the image. For an image pixel $ip$, we compare the surrounding patch $p_{lss}$ with size $5\times5$ and a larger surrounding region $r_{lss}$ with size $40\times40$. We convert the $p_{lss}$ and $r_{lss}$ into Lab color space and calculate the sum of square difference (SSD) in L channel. The result $SSD(p_{lss}, r_{lss})$ contains $8\times8$ values, denoting the differences between $p_{lss}$ and patches in $r_{lss}$ without overlap. Then, the local self-similarity of image pixel $ip$ is estimated by
	}
	\begin{equation}
		\label{eq:20}
		LSS(ip) = \max(\exp(-SSD(p_{lss}, r_{lss}))).
	\end{equation}
	{
		The $SSD(p_{lss}, r_{lss})$ describes the difference between the central patch and its neighbors. When $SSD(p_{lss}, r_{lss})$ is smaller, $LSS(ip)$ becomes larger and there is higher local self-similarity surrounding image pixel $ip$. As such, $LSS(\cdot)$ is a feasible descriptor for the local self-similarity.
	}
	
	{
		The local self-similarity of the image is estimated as follows. Firstly, we separate the image into several $40\times40$ patches without overlap. Then, we calculate the LSS for every patch. The average of LSS is regarded as the local self-similarity of the image.
	}
	
	{
		In general, the images contain fewer high-frequency details and more clear structural textures are with higher local self-similarity and more suitable for HSRNet to restore. Figure~\ref{fig:r2q2} shows the comparison between the local self-similarity and the restoration performance. The values under the image denote the PSNR/SSIM/LSS results. (a) row of the figure shows the images with lower LSS and (b) row shows the images with higher LSS. We can find that images in Figure~\ref{fig:r2q2} (a) contain more high-frequency details and tiny textures, which cause lower LSS values. The PSNR/SSIM results of these images are still lower. In contrast to the (a) row, images in Figure~\ref{fig:r2q2} (b) have more clear structural textures and fewer high-frequency details. In this point of view, the PSNR/SSIM results of the images are much higher with larger LSS values.
	}
	
	{
		From the perspective a user, LSS can be a good criterion to judge whether the image should be restored by HSRNet. The image with higher LSS value usually can be well reconstructed with higher PSNR results. To address this point, we calculate the Pearson's linear correlation coefficient~(PLCC) and the Spearman's rank correlation coefficient~(SRCC) to show the consistency between LSS and PSNR. Table~\ref{tab:r2q3} shows the PLCC/SRCC results between PSNR and LSS on different benchmarks with $\textbf{BI}\times4$. We can find that the correlation coefficients of HR and SR images are higher than 0.5 on all benchmarks. The higher PLCC/SRCC results denote higher correlations between LSS and PSNR. It should be noted that the correlations of LR are lower than SR and HR. This is because there is aliasing effect in the LR images, causing the mixture of textures and decrease the LSS. Even though, the PLCC/SRCC results of Urban100 and Manga109 are still higher than 0.5. HSRNet suppress the aliasing effect in SR images, and improves the PLCC/SRCC results. In this point of view, the user can calculate the LSS after HSRNet as a posterior. If the LSS is higher, then the restored image usually holds a better objective quality.
	}

	\section{Conclusion}
	In this paper, we proposed a hierarchical similarity learning network for aliasing suppressing image super-resolution, which is termed as HSRNet. We found there are correlations and self-similarities among the adjacent patches in the natural images, and devised the hierarchical exploration blocks (HEBs) to learn the local image prior. To obtain the relations of adjacent feature and focus more on the high-frequency information, multi-level spatial attention (MSA) was proposed to boost the restoration performance. Furthermore, we considered the issue in the optimization perspective and proposed a hierarchical similarity learning super-resolution network (HSRNet) based on the half-quadratic splitting strategy. Experimental results show our HSRNet achieves better performance than other works. When compared with other iterative works, HSRNet achieves competitive or better performance with much fewer parameters and lower computation complexity. Furthermore, the visualization comparisons show HSRNet can suppress the aliasing and restore the correct textures more effectively than other works.
	
	
	%



	
	
	\bibliographystyle{IEEEtran}
	\bibliography{sr, main}

\begin{thebibliography}{10}
\providecommand{\url}[1]{#1}
\csname url@samestyle\endcsname
\providecommand{\newblock}{\relax}
\providecommand{\bibinfo}[2]{#2}
\providecommand{\BIBentrySTDinterwordspacing}{\spaceskip=0pt\relax}
\providecommand{\BIBentryALTinterwordstretchfactor}{4}
\providecommand{\BIBentryALTinterwordspacing}{\spaceskip=\fontdimen2\font plus
\BIBentryALTinterwordstretchfactor\fontdimen3\font minus
  \fontdimen4\font\relax}
\providecommand{\BIBforeignlanguage}[2]{{%
\expandafter\ifx\csname l@#1\endcsname\relax
\typeout{** WARNING: IEEEtran.bst: No hyphenation pattern has been}%
\typeout{** loaded for the language `#1'. Using the pattern for}%
\typeout{** the default language instead.}%
\else
\language=\csname l@#1\endcsname
\fi
#2}}
\providecommand{\BIBdecl}{\relax}
\BIBdecl

\bibitem{sr_review_pami2020}
Z.~Wang, J.~Chen, and S.~C. Hoi, ``Deep learning for image super-resolution: A
  survey,'' \emph{IEEE Transactions on Pattern Analysis and Machine
  Intelligence (TPAMI)}, pp. 1--1, 2020.

\bibitem{nyquist_tit2009}
T.~Blumensath and M.~E. Davies, ``Sampling theorems for signals from the union
  of finite-dimensional linear subspaces,'' \emph{IEEE Transactions on
  Information Theory (TIT)}, vol.~55, no.~4, pp. 1872--1882, 2009.

\bibitem{srcnn_pami2016}
C.~Dong, C.~C. Loy, K.~He, and X.~Tang, ``Image super-resolution using deep
  convolutional networks,'' \emph{IEEE Transactions on Pattern Analysis and
  Machine Intelligence (TPAMI)}, vol.~38, no.~2, pp. 295--307, 2016.

\bibitem{vdsr_cvpr2016}
J.~Kim, J.~K. Lee, and K.~M. Lee, ``Accurate image super-resolution using very
  deep convolutional networks,'' in \emph{IEEE Conference on Computer Vision
  and Pattern Recognition (CVPR)}, 2016, pp. 1646--1654.

\bibitem{edsr_cvprw2017}
B.~Lim, S.~Son, H.~Kim, S.~Nah, and K.~M. Lee, ``Enhanced deep residual
  networks for single image super-resolution,'' in \emph{IEEE Conference on
  Computer Vision and Pattern Recognition Workshops (CVPRW)}, 2017, pp.
  1132--1140.

\bibitem{lapsrn_pami2019}
W.-S. Lai, J.-B. Huang, N.~Ahuja, and M.-H. Yang, ``Fast and accurate image
  super-resolution with deep laplacian pyramid networks,'' \emph{IEEE
  Transactions on Pattern Analysis and Machine Intelligence (TPAMI)}, vol.~41,
  no.~11, pp. 2599--2613, 2019.

\bibitem{landmarksr_tip2013}
H.~Yue, X.~Sun, J.~Yang, and F.~Wu, ``Landmark image super-resolution by
  retrieving web images,'' \emph{IEEE Transactions on Image Processing (TIP)},
  vol.~22, no.~12, pp. 4865--4878, 2013.

\bibitem{anchorsr_iccv2013}
R.~Timofte, V.~De, and L.~V. Gool, ``Anchored neighborhood regression for fast
  example-based super-resolution,'' in \emph{IEEE International Conference on
  Computer Vision (ICCV)}, 2013, pp. 1920--1927.

\bibitem{gssr_tmm2017}
J.~Liu, W.~Yang, X.~Zhang, and Z.~Guo, ``Retrieval compensated group structured
  sparsity for image super-resolution,'' \emph{IEEE Transactions on Multimedia
  (TMM)}, vol.~19, no.~2, pp. 302--316, 2017.

\bibitem{crossnet_eccv2018}
H.~Zheng, M.~Ji, H.~Wang, Y.~Liu, and L.~Fang, ``Crossnet: An end-to-end
  reference-based super resolution network using cross-scale warping,'' in
  \emph{European Conference on Computer Vision (ECCV)}, vol. 11210, 2018, pp.
  87--104.

\bibitem{srntt_cvpr2019}
Z.~Zhang, Z.~Wang, Z.~Lin, and H.~Qi, ``Image super-resolution by neural
  texture transfer,'' in \emph{IEEE/CVF Conference on Computer Vision and
  Pattern Recognition (CVPR)}, 2019, pp. 7974--7983.

\bibitem{msrn_eccv2018}
J.~Li, F.~Fang, K.~Mei, and G.~Zhang, ``Multi-scale residual network for image
  super-resolution,'' in \emph{European Conference on Computer Vision (ECCV)},
  2018, pp. 527--542.

\bibitem{urban100}
J.~{Huang}, A.~{Singh}, and N.~{Ahuja}, ``Single image super-resolution from
  transformed self-exemplars,'' in \emph{IEEE Conference on Computer Vision and
  Pattern Recognition (CVPR)}, 2015, pp. 5197--5206.

\bibitem{espcn_cvpr2016}
W.~Shi, J.~Caballero, F.~Huszár, J.~Totz, A.~P. Aitken, R.~Bishop,
  D.~Rueckert, and Z.~Wang, ``Real-time single image and video super-resolution
  using an efficient sub-pixel convolutional neural network,'' in \emph{IEEE
  Conference on Computer Vision and Pattern Recognition (CVPR)}, 2016, pp.
  1874--1883.

\bibitem{srdensenet_iccv2017}
T.~Tong, G.~Li, X.~Liu, and Q.~Gao, ``Image super-resolution using dense skip
  connections,'' in \emph{IEEE International Conference on Computer Vision
  (ICCV)}, 2017, pp. 4809--4817.

\bibitem{rdn_cvpr2018}
Y.~Zhang, Y.~Tian, Y.~Kong, B.~Zhong, and Y.~Fu, ``Residual dense network for
  image super-resolution,'' in \emph{IEEE/CVF Conference on Computer Vision and
  Pattern Recognition (CVPR)}, 2018, pp. 2472--2481.

\bibitem{rcan_eccv2018}
Y.~Zhang, K.~Li, K.~Li, L.~Wang, B.~Zhong, and Y.~Fu, ``Image super-resolution
  using very deep residual channel attention networks,'' in \emph{European
  Conference on Computer Vision (ECCV)}, 2018, pp. 294--310.

\bibitem{oisr_cvpr2019}
X.~He, Z.~Mo, P.~Wang, Y.~Liu, M.~Yang, and J.~Cheng, ``Ode-inspired network
  design for single image super-resolution,'' in \emph{IEEE/CVF Conference on
  Computer Vision and Pattern Recognition (CVPR)}, 2019, pp. 1732--1741.

\bibitem{dbpn_pami2020}
M.~Haris, G.~Shakhnarovich, and N.~Ukita, ``Deep back-projection networks for
  single image super-resolution,'' \emph{IEEE Transactions on Pattern Analysis
  and Machine Intelligence (TPAMI)}, pp. 1--1, 2020.

\bibitem{srfbn_cvpr2019}
Z.~Li, J.~Yang, Z.~Liu, X.~Yang, G.~Jeon, and W.~Wu, ``Feedback network for
  image super-resolution,'' in \emph{IEEE/CVF Conference on Computer Vision and
  Pattern Recognition (CVPR)}, 2019, pp. 3862--3871.

\bibitem{dan_cvpr2020}
Y.~Guo, J.~Chen, J.~Wang, Q.~Chen, J.~Cao, Z.~Deng, Y.~Xu, and M.~Tan,
  ``Closed-loop matters: Dual regression networks for single image
  super-resolution,'' in \emph{IEEE/CVF Conference on Computer Vision and
  Pattern Recognition (CVPR)}, 2020, pp. 5406--5415.

\bibitem{csnl_cvpr2020}
Y.~Mei, Y.~Fan, Y.~Zhou, L.~Huang, T.~S. Huang, and H.~Shi, ``Image
  super-resolution with cross-scale non-local attention and exhaustive
  self-exemplars mining,'' in \emph{IEEE/CVF Conference on Computer Vision and
  Pattern Recognition (CVPR)}, 2020, pp. 5689--5698.

\bibitem{carn_eccv2018}
N.~Ahn, B.~Kang, and K.~Sohn, ``Fast, accurate, and lightweight
  super-resolution with cascading residual network,'' in \emph{European
  Conference on Computer Vision (ECCV)}, 2018, pp. 256--272.

\bibitem{idn_cvpr2018}
Z.~Hui, X.~Wang, and X.~Gao, ``Fast and accurate single image super-resolution
  via information distillation network,'' in \emph{IEEE/CVF Conference on
  Computer Vision and Pattern Recognition (CVPR)}, 2018, pp. 723--731.

\bibitem{imdn_mm2019}
Z.~Hui, X.~Gao, Y.~Yang, and X.~Wang, ``Lightweight image super-resolution with
  information multi-distillation network,'' in \emph{ACM International
  Conference on Multimedia (MM)}, 2019, p. 2024–2032.

\bibitem{aim2020}
K.~Zhang, M.~Danelljan, Y.~Li, and et. al., ``{AIM} 2020 challenge on efficient
  super-resolution: Methods and results,'' in \emph{European Conference on
  Computer Vision Workshops (ECCVW)}, 2020, pp. 5--40.

\bibitem{senet_cvpr2018}
J.~Hu, L.~Shen, and G.~Sun, ``Squeeze-and-excitation networks,'' in
  \emph{IEEE/CVF Conference on Computer Vision and Pattern Recognition (CVPR)},
  2018, pp. 7132--7141.

\bibitem{ircnn_cvpr2017}
K.~Zhang, W.~Zuo, S.~Gu, and L.~Zhang, ``Learning deep cnn denoiser prior for
  image restoration,'' in \emph{IEEE Conference on Computer Vision and Pattern
  Recognition (CVPR)}, 2017, pp. 2808--2817.

\bibitem{dpsr_cvpr2019}
K.~Zhang, W.~Zuo, and L.~Zhang, ``Deep plug-and-play super-resolution for
  arbitrary blur kernels,'' in \emph{IEEE/CVF Conference on Computer Vision and
  Pattern Recognition (CVPR)}, 2019, pp. 1671--1681.

\bibitem{usrnet_cvpr2020}
K.~Zhang, L.~Van~Gool, and R.~Timofte, ``Deep unfolding network for image
  super-resolution,'' in \emph{IEEE/CVF Conference on Computer Vision and
  Pattern Recognition (CVPR)}, 2020, pp. 3214--3223.

\bibitem{isrn_tmm2021}
Y.~Liu, S.~Wang, J.~Zhang, S.~Wang, S.~Ma, and W.~Gao, ``Iterative network for
  image super-resolution,'' \emph{IEEE Transactions on Multimedia}, pp. 1--1,
  2021.

\bibitem{plug_and_play_admm_tci2017}
S.~H. Chan, X.~Wang, and O.~A. Elgendy, ``Plug-and-play admm for image
  restoration: Fixed-point convergence and applications,'' \emph{IEEE
  Transactions on Computational Imaging (TCI)}, vol.~3, no.~1, pp. 84--98,
  2017.

\bibitem{jmdl_tip2020}
X.~Deng and P.~L. Dragotti, ``Deep coupled ista network for multi-modal image
  super-resolution,'' \emph{IEEE Transactions on Image Processing (TIP)},
  vol.~29, pp. 1683--1698, 2020.

\bibitem{istanet_cvpr2018}
J.~Zhang and B.~Ghanem, ``Ista-net: Interpretable optimization-inspired deep
  network for image compressive sensing,'' in \emph{IEEE/CVF Conference on
  Computer Vision and Pattern Recognition (CVPR)}, 2018, pp. 1828--1837.

\bibitem{admmnet_iccv2019}
J.~Ma, X.-Y. Liu, Z.~Shou, and X.~Yuan, ``Deep tensor admm-net for snapshot
  compressive imaging,'' in \emph{2019 IEEE/CVF International Conference on
  Computer Vision (ICCV)}, 2019, pp. 10\,222--10\,231.

\bibitem{div2k}
E.~{Agustsson} and R.~{Timofte}, ``Ntire 2017 challenge on single image
  super-resolution: Dataset and study,'' in \emph{IEEE Conference on Computer
  Vision and Pattern Recognition Workshops (CVPRW)}, 2017, pp. 1122--1131.

\bibitem{set5}
M.~Bevilacqua, A.~Roumy, C.~Guillemot, and M.~line Alberi~Morel,
  ``Low-complexity single-image super-resolution based on nonnegative neighbor
  embedding,'' in \emph{British Machine Vision Conference (BMVC)}, 2012, pp.
  135.1--135.10.

\bibitem{set14}
R.~Zeyde, M.~Elad, and M.~Protter, ``On single image scale-up using
  sparse-representations,'' in \emph{International Conference on Curves and
  Surfaces}.\hskip 1em plus 0.5em minus 0.4em\relax Springer, 2010, pp.
  711--730.

\bibitem{b100}
D.~{Martin}, C.~{Fowlkes}, D.~{Tal}, and J.~{Malik}, ``A database of human
  segmented natural images and its application to evaluating segmentation
  algorithms and measuring ecological statistics,'' in \emph{IEEE International
  Conference on Computer Vision (ICCV)}, vol.~2, 2001, pp. 416--423 vol.2.

\bibitem{manga109}
Y.~Matsui, K.~Ito, Y.~Aramaki, A.~Fujimoto, T.~Ogawa, T.~Yamasaki, and
  K.~Aizawa, ``Sketch-based manga retrieval using manga109 dataset,''
  \emph{Multimedia Tools and Applications (MTA)}, vol.~76, no.~20, pp.
  21\,811--21\,838, 2017.

\bibitem{fsrcnn_eccv2016}
C.~Dong, C.~C. Loy, and X.~Tang, ``Accelerating the super-resolution
  convolutional neural network,'' in \emph{European Conference on Computer
  Vision (ECCV)}, 2016, pp. 391--407.

\bibitem{drcn_cvpr2016}
J.~Kim, J.~K. Lee, and K.~M. Lee, ``Deeply-recursive convolutional network for
  image super-resolution,'' in \emph{IEEE Conference on Computer Vision and
  Pattern Recognition (CVPR)}, 2016, pp. 1637--1645.

\bibitem{drrn_cvpr2017}
Y.~Tai, J.~Yang, and X.~Liu, ``Image super-resolution via deep recursive
  residual network,'' in \emph{IEEE Conference on Computer Vision and Pattern
  Recognition (CVPR)}, 2017, pp. 2790--2798.

\bibitem{memnet_iccv2017}
Y.~Tai, J.~Yang, X.~Liu, and C.~Xu, ``Memnet: A persistent memory network for
  image restoration,'' in \emph{IEEE International Conference on Computer
  Vision (ICCV)}, 2017, pp. 4549--4557.

\bibitem{degree_tip2017}
W.~{Yang}, J.~{Feng}, J.~{Yang}, F.~{Zhao}, J.~{Liu}, Z.~{Guo}, and S.~{Yan},
  ``Deep edge guided recurrent residual learning for image super-resolution,''
  \emph{IEEE Transactions on Image Processing}, vol.~26, no.~12, pp.
  5895--5907, 2017.

\bibitem{ran_csvt2019}
Y.~Wang, L.~Wang, H.~Wang, and P.~Li, ``Resolution-aware network for image
  super-resolution,'' \emph{{IEEE} Transactions on Circuits and Systems for
  Video Technology}, vol.~29, no.~5, pp. 1259--1269, 2019.

\bibitem{dncl_csvt2019}
C.~Xie, W.~Zeng, and X.~Lu, ``Fast single-image super-resolution via deep
  network with component learning,'' \emph{{IEEE} Transactions on Circuits and
  Systems for Video Technology}, vol.~29, no.~12, pp. 3473--3486, 2019.

\bibitem{filternet_csvt2020}
F.~Li, H.~Bai, and Y.~Zhao, ``Filternet: Adaptive information filtering network
  for accurate and fast image super-resolution,'' \emph{{IEEE} Transactions on
  Circuits and Systems for Video Technology}, vol.~30, no.~6, pp. 1511--1523,
  2020.

\bibitem{mrfn_tmm2020}
Z.~He, Y.~Cao, L.~Du, B.~Xu, J.~Yang, Y.~Cao, S.~Tang, and Y.~Zhuang, ``{MRFN:}
  multi-receptive-field network for fast and accurate single image
  super-resolution,'' \emph{{IEEE} Transactions on Multimedia}, vol.~22, no.~4,
  pp. 1042--1054, 2020.

\bibitem{seanet_tip2020}
F.~{Fang}, J.~{Li}, and T.~{Zeng}, ``Soft-edge assisted network for single
  image super-resolution,'' \emph{IEEE Transactions on Image Processing},
  vol.~29, pp. 4656--4668, 2020.

\bibitem{s_LWSR_tip2020}
B.~{Li}, B.~{Wang}, J.~{Liu}, Z.~{Qi}, and Y.~{Shi}, ``s-lwsr: Super
  lightweight super-resolution network,'' \emph{IEEE Transactions on Image
  Processing (TIP)}, vol.~29, pp. 8368--8380, 2020.

\bibitem{crosssrn}
Y.~Liu, Q.~Jia, X.~Fan, S.~Wang, S.~Ma, and W.~Gao, ``Cross-srn:
  Structure-preserving super-resolution network with cross convolution,''
  \emph{IEEE Transactions on Circuits and Systems for Video Technology}, pp.
  1--1, 2021.

\bibitem{lss}
E.~Shechtman and M.~Irani, ``Matching local self-similarities across images and
  videos,'' in \emph{IEEE Conference on Computer Vision and Pattern Recognition
  (CVPR)}, 2007, pp. 1--8.

\end{thebibliography}
\end{document}